\documentclass[lettersize,journal]{IEEEtran}
\usepackage{amsmath,amsfonts}
\usepackage{algorithmic}
\usepackage{algorithm}
\usepackage{array}
\usepackage[caption=false,font=normalsize,labelfont=sf,textfont=sf]{subfig}
\usepackage{textcomp}
\usepackage{stfloats}
\usepackage{url}
\usepackage{verbatim}
\usepackage{graphicx}
\usepackage{cite}
\usepackage{mathrsfs}
\usepackage{amssymb}
\usepackage{footnote}
\usepackage{multirow}
\usepackage{tabularx}
\usepackage{booktabs}
\usepackage{utfsym}
\usepackage{bm}
\usepackage{mathrsfs}
\begin{document}

\title{Artificial-Spiking Hierarchical Networks for Vision-Language Representation Learning}

\author{Yeming~Chen,
        Siyu~Zhang$^{\ast}$,
        Yaoru~Sun$^{\ast}$,
        Weijian~Liang,
       and~Haoran~Wang
\thanks{Yeming Chen, Siyu Zhang, Yaoru Sun, Weijian Liang, and Haoran Wang are with the Department of Computer Science and Technology, Tongji University, Shanghai 201804, China (e-mail:2130769@tongji.edu.cn; zsyzsy@tongji.edu.cn; yaoru@tongji.edu.cn; liangweijian@tongji.edu.cn; wanghaoran\_tj@tongji.edu.cn).}
\thanks{$\ast$ Corresponding author.}}

\markboth{}%
{Shell \MakeLowercase{\textit{et al.}}: A Sample Article Using IEEEtran.cls for IEEE Journals}


\maketitle

\begin{abstract}
With the success of self-supervised learning, multimodal foundation models have rapidly adapted a wide range of downstream tasks driven by vision and language (VL) pre-training. State-of-the-art methods achieve impressive performance by pre-training on large-scale datasets. However, bridging the semantic gap between the two modalities remains a non-negligible challenge for VL tasks. In this work, we propose an efficient computation framework for multimodal alignment by introducing a novel visual semantic module to further improve the performance of the VL tasks. Specifically, we propose a flexible model, namely Artificial-Spiking Hierarchical Networks (ASH-Nets), which combines the complementary advantages of Artificial neural networks (ANNs) and Spiking neural networks (SNNs) to enrich visual semantic representations. In particular, a visual concrete encoder and a semantic abstract encoder are constructed to learn continuous and discrete latent variables to enhance the flexibility of semantic encoding. Considering the spatio-temporal properties of SNNs modeling, we introduce a contrastive learning method to optimize the inputs of similar samples. This can improve the computational efficiency of the hierarchical network, while the augmentation of hard samples is beneficial to the learning of visual representations. Furthermore, the Spiking to Text Uni-Alignment Learning (STUA) pre-training method is proposed, which only relies on text features to enhance the encoding ability of abstract semantics. We validate the performance on multiple well-established downstream VL tasks. Experiments show that the proposed ASH-Nets achieve competitive results.

\end{abstract}


\section{Introduction}
\IEEEPARstart{V}{ision} and language are two indispensable significant systems in human cognition. Within the field of multimodality, there exists a large amount of research work on VL tasks. Many VL methods are designed for specific tasks, which encompasses Visual question answering (VQA) \cite{ref1}, image-text retrieval \cite{ref2}, natural language for visual reasoning (NLVR) \cite{ref3}, etc. These tasks require VL pre-training models to jointly encode and understand these two different modalities.
A good visual representation is critical for VL models. Early visual encoding methods focused on two-stage training pipeline, which exploited the Convolutional Neural Networks (CNNs) \cite{ref4} and Recurrent Neural Networks (RNNs) \cite{ref5}, and combined the VL features to reason the corresponding answer. With the proposal of bottom-up and top-down attention model \cite{ref6}, object detection (e.g., Faster R-CNN \cite{ref7}) as a collection of bounding boxes or region-based features extraction method has been widely used to pre-trained on the Visual Genome (VG) dataset \cite{ref8}. However, visual features based on salient regions usually represent parts of an image. In addition, spatial relations between overlapping objects are ignored, which leads to insufficient understanding of visual semantic information. Therefore, it is not optimal for specific question tasks.

In recent years, some works have revisited grid-based features to learn effective visual representations. For example, Jiang \textit{et al.} \cite{ref9} proposed a grid-based VQA model that can achieve equally performant region features. Furthermore, they emphasize that the visual semantic representation is more important than the feature “format” itself (i.e., grid or region). As we have observed, the early VL methods have experienced from exploring global features based on CNN visual grids to object detection-based local features, and back again. With further developments, self-supervised pre-training techniques are becoming more popular since their strong ability to deal with data-hungry issues \cite{ref10}. Recent research has applied different VL pre-training methods, especially for multi-modal learning in a large-scale end-to-end manner. For instance, Huang et al. \cite{ref11} designed a CNN-based grid features VL pre-training model that learned concrete yet compact visual representations by using a vision dictionary. More recently, vision transformers (ViTs) as a hot topic inspired by transformers. Some ViT-based methods \cite{ref12},\cite{ref13} have been proposed for model pre-training on large-scale image data. Although these works have made significant progress in overcoming the limitations of region-based image features. However, there are still some serious challenges in multi-modal alignment tasks. On the one hand, the input to the vision transformer has no pre-existing vocabulary, which is different from the discrete signal spaces (e.g., words or sub-word units) of the tokenized vocabularies built for languages. On the other hand, long visual sequences at pixel-level will lead to computational inefficiency and expensive.

\IEEEpubidadjcol
To overcome the above problems, we propose a novel end-to-end multimodal alignment model. Unlike previous methods, our model aims to encode visual basis vectors by optimizing image patches. To this end, we designed an efficient and adaptive visual semantic module. We encode input images as grids of patches that are trained by learnable Artificial-Spiking Hierarchical Networks (ASH-Nets) for high-level visual semantic features. Since the raw visual signal is in a continuous high-dimensional space, which leads to signal spaces difference in the process of aligning with language. Therefore, the ASH-Nets model is more suitable for building an efficient visual semantic encoder due to its sparsity, flexibility, and powerful hierarchical computing capabilities. Furthermore, we integrate input samples with similar semantics via a contrastive learning method. On the one hand, the membrane potential reset caused by a single input sample is avoided, which greatly improves the learning efficiency of SNNs. On the other hand, hard samples generated by similar samples recombination can speed up the model convergence and improve accuracy. In addition following the previous pre-training tasks, such as Image-Text Matching (ITM) \cite{ref14}, Masked Language Modeling (MLM)\cite{ref15} and Masked Visual Modeling (MVM), we also design Spiking to Text Uni-Alignment Learning (STUA), which is used to align multi-modal semantic labels for pre-training tasks. The pre-training experiments of the proposed ASH-Nets model are conducted on typical public datasets, including Visual Genome (VG) and MSCOCO \cite{ref16}. In contrast to the recent VL pre-training methods, we achieve solid performance on multiple popular VL tasks (e.g., VQA). Additionally, we perform ablation studies to demonstrate that the visual semantics module designed based on our model is simpler and easier to implement during training. 
The main contributions of this work are summarized as follows.
\begin{itemize}[\IEEEsetlabelwidth{Z}]
\item We propose ASH-Nets, which combine ANNs-based visual concrete encoder and SNNs-based semantic abstract encoder to learn the underlying visual semantic features composition, relational, and hierarchical structures of inter-modality. ASH-Nets not only provide the intrinsic capability to process complex signals but also bridge the semantic gap between different modalities, thereby reducing the alignment difficulty.
\item  We design a contrastive learning method to optimize hierarchical network inputs. By fully considering the spatio-temporal properties of SNNs, this method enhances the robustness of semantic representation while minimizing computational overhead. 
\item We propose a novel Spiking to Text Uni-Alignment Learning (STUA) for pre-training tasks. STUA can encourage the model to improve the learning ability of abstract semantics from text information. 
\item Our proposed method is effectively validated on four well-established downstream VL tasks and achieves better performance.
\end{itemize}

The remainder of this work is organized as follows. Section II introduces the related work of VL. Section III mainly details the designed ASH-Nets method. In Section IV, the experimental results on four downstream tasks and the comparisons with recent state-of-the-art methods are listed. Finally, discussions and conclusions are given in Section V.

\section{Related Work}
\subsection{Visual encoder embedding} 
Early visual representation mainly involved two stages: \textit{i)} plain CNN. \textit{ii)} object detector. Benefiting from the successful application of CNN on image classification, early works directly used CNN pre-trained on ImageNet as a classification model to encode visual information. Yang \textit{et al}. \cite{ref17} designed an efficient stack attention network for VQA, which used multiple times image queries to reason relevant answers. Shih \textit{et al}. \cite{ref18} first focused on the idea of region features and applied it to VQA. The encoding way of visual features based on local regions has attracted the attention of researchers. In particular, bounding boxes have been widely optimized and processed due to the simplicity of representing object locations. Anderson \textit{et al}. \cite{ref6} developed a combined bottom-up and top-down attention (BUTD) model pre-trained on the VG, where region-based salient features were encoded using an object detection method. Region-based features have soon dominated the design of many VL works, such as VQA and image captioning \cite{ref19}, \cite{ref20}. However, some important visual information is limited to the given region categories and makes it difficult to understand much broader semantic relations. Recently, some works attempt to bridge the VL modality gap. For instance, Zhou \textit{et al}. \cite{ref21} introduced bidirectional attention and sequence-to-sequence mask to construct a unified pre-training method. Li \textit{et al}. \cite{ref22} developed a unified mPLUG model, which contained cross-modal skip-connections to improve multimodal alignment and time-consuming issues. Lu \textit{et al}. \cite{ref23} proposed effective multi-task learning to train a single model on 12 popular datasets. It is worth noting that the annotation of salient visual regions is obtained from noun or gerund phrases such as text labels. Hence, the language generally contains more semantic information than salient visual regions. This asymmetry poses serious challenges to effective fusion between multi-modalities. 

\subsection{Pre-training mechanism} 
Transformer has successfully replaced RNN owing to its strong semantic capture ability and higher computational efficiency dominating the current research domain. Many different methods have been devised to pre-train VL models in an end-to-end fashion. Specifically, existing methods can be classified into CNN-based grid features and ViT-based patch features based on their encoding manner. Huang \textit{et al}. \cite{ref24} designed a unified end-to-end Pixel-BERT model, which aligned semantic connections in image pixels and text level by deep multi-modal transformers. Shen \textit{et al}. \cite{ref25} proposed to use contrastive language-image pre-training as the visual encoder and applied it to downstream tasks. More recently, Dosovitskiy \textit{et al}. \cite{ref26} developed vision transformers, where each image was split into image patches to meet the input requirement of the transformer. Later on, Kim \textit{et al}. \cite{ref12} presented a minimal vision-and-language transformer (ViLT), and the model was pre-trained on image-text datasets. Although different ViT-based pre-training methods achieved higher performance driven by large-scale datasets. Both the heavy-weight structure and expensive computation cost make the model difficult to train. 

Different from various ViT variants for VL pre-training, the key idea of our work is to build visual basis vectors, aiming at optimizing pixel-level long visual sequences to enhance semantic representation. Since raw visual signals are in a continuous high-dimensional space, visual representations are more detailed and diverse than language features. To this end, we present a learnable and designable visual semantic encoder based on the ASH-Nets to address multi-modal information asymmetry. Furthermore, the proposed ASH-Nets can flexibly and dynamically capture rich contextual semantic information to better support downstream VL tasks. 

\section{Methods}
In this section, we first introduce a multi-network hierarchical model structure by combining trainable continuous units and discrete units as the learnable visual representation, followed by the details of the visual encoders for different distribution states. 

\subsection{Hybrid modulation structure}
Existing methods generally use ANN-based workloads and evaluation metrics to validate SNN models. However, the original datasets for ANNs are static images, which prevents the spatio-temporal advantages of SNNs from being fully utilized. Therefore, there are two major tasks to be implemented: \textit{i)} design suitable workloads and optimize inputs and outputs for SNNs and \textit{ii)} construct ANNs-SNNs hierarchical multi-network structure.

\begin{figure*}[t!]
\begin{center}
\includegraphics[width=6in]{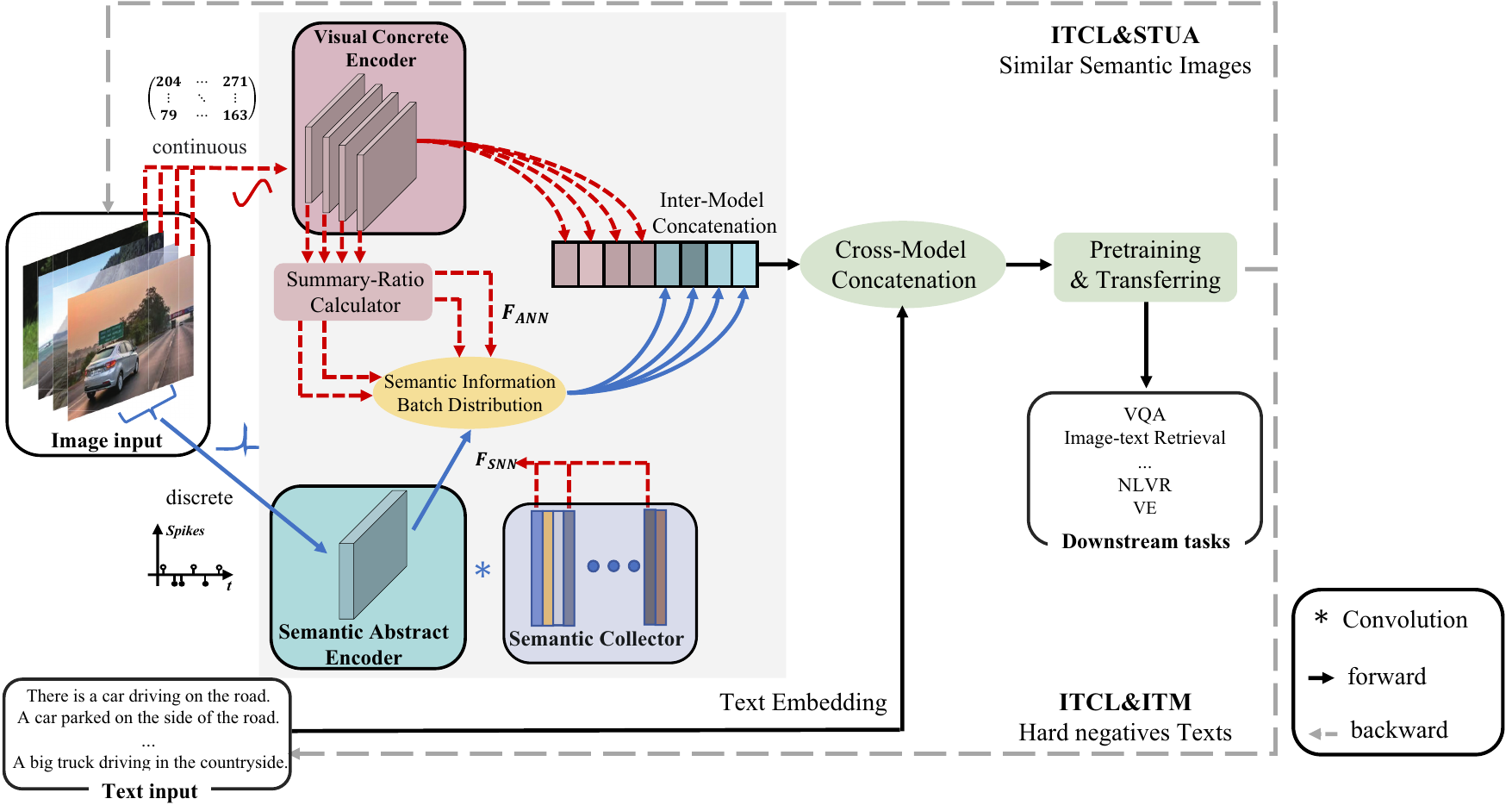}
\end{center}
   \caption{The architecture of the proposed ASH-Nets. For the vision module, we proposed an ANNs-based visual concrete encoder (dashed red lines transmission path) and an SNNs-based semantic abstract encoder (solid blue lines transmission path) to learn visual semantic representations. To align semantic features, we use multi-layer transformers to concatenate the output of images and language text.}
\label{fig_1}
\end{figure*}
The advantage of SNN is that it can simultaneously process spatial information and historical information in the temporal dimension. This also indirectly shows that the neurons of SNNs cannot be reset for each image. To avoid the computational inefficiency caused by the evolution of SNNs into a complex version of ANNs, we need to design reasonable evaluation criteria for the outputs of SNNs, instead of implementing downstream tasks such as accuracy prediction. Additionally, SNNs possess sparse features and low-power attributes, we argue that it is feasible to combine ANNs and SNNs to construct interwoven and complementary multi-network model with a dynamic ratio “fusion”. On the one hand, the strengths of ANNs and SNNs can be greatly exerted, and the complementarity in the form of information representation is realized. On the other hand, the implementation of dynamic fusion attributes provides the model itself with a high degree of freedom, as well as the inherent ability to process complex spatiotemporal information. In Fig. \ref{fig_1}, we provide the overall architecture of the designed end-to-end ASH-Nets pre-training model. Specifically, we show the key components of the visual embedding mechanism, including continuous signals parsed by visual concrete encoders and discrete signals parsed by semantic abstract encoders.
\begin{figure}[t!]
\begin{center}
\includegraphics[width=2in]{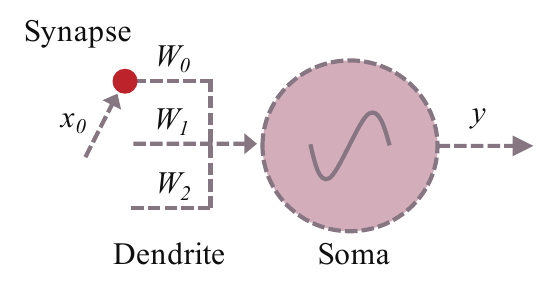}\\
\includegraphics[width=2in]{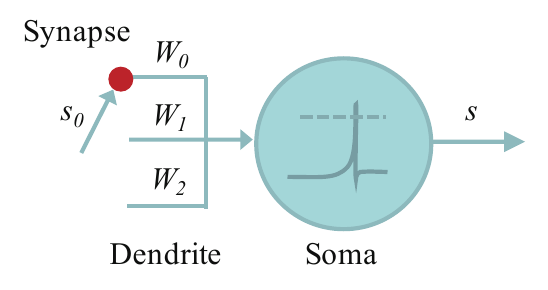}
\end{center}
   \caption{Basic neuron model: (a) synchronous-continuous signals of ANNs, (b) asynchronous-discrete signals of SNNs.}
\label{fig_2}
\end{figure}

\textit{1) Details of neurons:} Artificial neural networks (ANNs) have brought amazing successes via mature models, open-source datasets, various benchmarks, and powerful computing platforms. However, issues such as power consumption and computational overhead increase significantly with ever-increasing network size. Recently, SNNs have been widely studied in the computer vision field. Unlike conventional ANNs, SNNs can perform information transformation and propagation on the discrete spike trains with low energy cost by mimicking the neuronal dynamics of the brain. Fig. \ref{fig_2} shows the state of a single neuron of ANN and SNN, respectively. Specifically, ANNs comprise neurons interconnected with synapses, where neurons propagate information using high-precision and real values encoded activations. The basic ANN neuron is expressed as follows:
\begin{equation}
\label{eq:1}  
y = \sigma(w_{i}\ast x_{i}+b)+\delta(x_{i})  
\end{equation}
where $\sigma(\cdot)$ is the nonlinear activation function. $\ast$ represents convolution operation. $x_{i}$ and $y$ represent the linear inputs and outputs, respectively. $w_{i}$  and $b$ represent the synaptic weight and bias parameters, respectively. $i$ is the index of the input neuron. $\delta$ represents up-sampled data, which uses alignment to facilitate residual connections. By contrast, SNNs transmit information through a series of discrete asynchronous spikes rather than continuous values. Spike timing dependent plasticity (STDP) \cite{ref27} training algorithm is employed to enhance the application accuracy of SNNs. To enable spiking dynamics with gradient descent, we select the popular leaky-integrate-and-fire LIF \cite{ref28} neurons to model, the expression is defined as: 
\begin{equation}
\label{eq:2}  
s = argmax(U_{r2}, (u_{t}-U_{r1}+\sum\nolimits_{k=0} ^{t}K(t,x_{t}))) 
\end{equation}
where $U_{r1}$ and $U_{r2}$ represent the resting potential and reset potential, respectively. $u_{t}$ represents the membrane potential at the current time step $t$. $K(\cdot)$ represents a kernel function that describes the time step decay from 0 to t. When the membrane potential $u$ exceeds the threshold $U_{r2}$, $s=1$, otherwise $u$ will be reset at the moment, $s=0$. The corresponding soft-activated expression is given by
\begin{equation}
\label{eq:3}  
s = (u_{t}-U_{r1}+\sum\nolimits_{k=0} ^{t}K(t,x_{t}))-U_{r2}
\end{equation}

Obviously, directly using the impulse function for gradient descent will make the training of the network unstable. Therefore, many surrogate gradient methods \cite{ref29} have been proposed recently. Here, we choose the tanh function for the surrogate function of the spiking neurons.

Compared to ANNs with spatial propagation and continuous activations, SNNs benefit from continuous spatio-temporal dynamics and event-driven spike communication paradigms, which are more temporally versatile. Different from the real values information representation used by ANNs, SNNs transmit information in binary spike trains (\textit{i.e.}, 0 or 1). It is worth noting that when the integration time window is equal to 1, the expensive multiplication between the input and weight can be removed. On this basis, SNNs can achieve low power consumption. Furthermore, there are also some similarities between ANN and SNN neurons, leaving room for deploying dynamic multi-network model.

\textit{2) Details of Visual Encoder:} In our proposed model, images with similar features are adopted as input data instead of being randomly sampled from the datasets. We introduce the image-text contrastive learning (ITC) \cite{ref30} pre-training task to complete the recall of similar samples matched by the two modalities. Specifically, we first need to freeze part of the SNNs model parameters during pre-training, and the remaining three to five epochs are iterated. Then, the trained SNNs model is unfrozen, and the sequence of input images is reorganized by using ITM. We assume that a batch of original image is $Pic_{k}, (k=1, \cdots, \varphi)$, and the updated batch is defined as $Retrieve(Pic_{k}, \varphi$). To integrate images with similar semantics, the function $Retrieve$ can be used to recall the $\varphi$ samples with the highest matching scores for the ITM in the dataset. Finally, we adaptively adjust the number of epochs to ensure that the overall training calculation does not increase (i.e., only change the order of the input samples). The design of the model input mainly contains the following two advantages: \textit{i)} The spatio-temporal modeling capabilities of SNNs can be fully utilized. We can model the common semantics of similar images in the time domain. In the spatial domain, it can model the abstract semantics of the image itself. Furthermore, SNNs can be reset after computing all images in a batch instead of after computing each image to improve computational efficiency. \textit{ii)} Image reorganization can be viewed as a hard sample sampling process, which helps model training alignment to learn more valuable representation information.

For the visual concrete encoder, each input image is performed independently. This indicates that there is no correlation between the samples. Even though the loss function is summed, the whole feature extraction process is trained independently. In this work, we formularize $V(\cdot)$ as a trainable CNNs concrete encoder, which takes image I of each batch size b as input and produces the visual features $F_{CNN}$. 

\begin{equation}
\label{eq:4}  
F_{CNN,b}=V(I_{b}; \Theta)
\end{equation}
where $I\in h\times w\times[256,256,3]\times b$, $F\in N\times M_1\times b$. $h$ and $w$ denote the height and width of the image, respectively. $N$ is the number of extracted feature vectors, and $M_1$ is the feature dimension. $\Theta$ is the parameter set of each batch in the CNN model.

For the semantic abstract encoder, the signal received by the spiking neuron is transformed from a single static image to a dynamic form when similar semantic visual information flows are formed. In particular, the information on each location needs to be fixed when ANNs process synchronous information. While SNNs need to respond to events when processing asynchronous information. Here, we convert images into spike patterns using a prevalent strategy known as probabilistic sampling. In terms of the neuron training method, threshold-dependent batch normalization (tdBN) \cite{ref31} is used to assist training, which helps to deepen the number of SNNs network layers and better extract abstract semantic information. The semantic abstract encoder $E(\cdot)$ is given by:

\begin{equation}
\label{eq:5}  
S_b=AC_{i=1}^b\mid E(P_{binary,i}; \gamma_{i})
\end{equation}
where $P_{binary}\in h\times w\times \{0,1\} \times T\times b$, $S_{b}\in N\times M_{2}\times b$. SNNs perform accumulate (AC) operations because spike events are binary values $\{0,1\}$ whose input Pbinary, $i$ is accumulated from $i=1$ to $b$ into a membrane potential. $T$ is the time window size in which the image is converted from $2D$ space to $1D$ spike signal. $\gamma$ is the parameter set. $M_2$ is the feature dimension of abstract semantics. Collectively, we choose similar images as the temporal dimension, which are continuously learned as event-stimulus spikes to bridge the semantic gap between multi-modalities.\\
Although SNNs closely mimic the behaviors of biological neural circuits, it is still not able to compete with ANNs in classification tasks at present. For some specific tasks, we consider that SNNs are more suitable for selecting “\textit{Yes or No}” rather than the question “\textit{What}”. Therefore, we set the SNNs as a discriminator to judge the existence of basic semantics by activating the visual information flows. And basic semantics are stored in the semantic collector. Here, we provide a visual collector $C\in R^{M_{1}\times M_2}$, which is inspired by SOHO \cite{ref11}. Unlike SOHO, we construct a high-degree-of-freedom semantic space through the SNNs model, and the basic semantics can be combined into abundant abstract deep-level features. The defined semantic collector can be expressed as:

\begin{equation}
\label{eq:6}  
F_{SNN,b}=\frac{\sum\nolimits_{j=1}^{M_2}C_{j}\ast S_b} {\sum\nolimits_{i=1}^Ns_{b,i}}
\end{equation}
where $F_{SNN,b}\in N\times M_1\times b$. $C_j$ represents the $j$th abstract semantic feature. 

To demonstrate the design value of the proposed model, we provide a performance comparison between ANNs and ANNs-SNNs in terms of computational efficiency. (\textit{See ablation study}) \\

\textit{3) Information transmission:} Since the images within each batch are input as a sequence of SNNs, the images within each batch have the same response on the SNNs. However, extracting the same abstract semantic information for each image is not conducive to model development. Given the difference in input behavior between ANNs and SNNs, the outputs of ANNs and SNNs need to be concatenated for inter-modal fusion. The summary ratio (SR) is used to redistribute abstract semantic information to measure the subtle differences between similar images. Therefore, we select $SR$ to coordinate the synchronization, time scale, and encoding schemes between ANNs and SNNs. Furthermore, model parameter configuration can be trained according to domain knowledge or learned to adapt. The output of the proposed ASH-Nets model can be formalized as: 

\begin{equation}
\label{eq:7}  
SR_b=\sigma\cdot\phi(F_{CNN,b})
\end{equation}
\begin{equation}
\label{eq:8}  
Output_b=SR_{b}\ast F_{SNN,b}+F_{CNN,b}
\end{equation}
where $\sigma(\cdot)$ represents the sigmoid function. $\phi$ is a fully-connected (FC) function.
Different from homogeneous information in ANNs or SNNs, our proposed method leverages the advantages of ANNs and SNNs with the support of a summary ratio. More importantly, our design provides flexible configurations and rich encoding schemes for hierarchical network model. Finally, VL features are implemented through multi-modal fusion to finish multimodal alignment. The relative trainable $SR$ in the ablation experiment is evaluated in Table \ref{table_11}.

\subsection{Alignmnet Transformers}
A single-stream transformer is proposed to model the VL modality in this work, which aims to learn a unified joint representation. Image and text tokens are concatenated and fed into a multi-layer transformer for modality fusion. We calculate the reverse gradient according to the custom training task to assist the model in learning semantic features of VL modalities.

\textit{1) Tokens Embedding:} 
For text embedding, we follow BERT \cite{ref32} to tokenize input sentences. The text embeddings are construct from word, position and token type embeddings. For image representation, the $F_{SNN, b}$ and $F_{CNN, b}$ features are obtained by the $Output_{b}$ of our ASH-Nets model can be represented as $M\times N$ embeddings. At the same time, these different output patches are regarded as different words, and the position of patches is encoded in the same manner as BERT. After the above operations, we can finally get the token embeddings and token masks of image and text alignment.

\textit{2) Label Generation:} 
Meaningful semantic information needs to be learned from the dataset to enable subsequent self-supervised training of the model. Such as MLM pre-training task, generally masks a certain word, which is used to predict other words. To evaluate the effectiveness of the proposed ASH-Nets model, we perform several downstream VL tasks. Our goal is to train SNNs and ANNs with continuous iterations on modality alignment. Every time a patch is masked, the response activation state of SNNs neurons will generate a multi-class label, which is designed for the subsequent STUA pre-training task. It is worth noting that we only use encoded text features by the transformer to predict, focusing on aligning abstract visual semantic features with text semantic features.

\subsection{Transformer Pre-training}
In order to learn multimodal representations for VL tasks, a multi-layer transformer is used that are pre-trained in medium-scale settings. We perform four existing pre-training tasks (ITM, MLM, Image-Text Contrastive Learning (ITC), and Masked Visual Modeling (MVM)), which are optimized jointly. Besides, we develop an efficient Spiking to Text Uni-Aligning Learning (STUA).\\

\textit{1) Image-Text Contrastive Learning (ITC):} We employ the ITC task from the unimodal encoders, which serves as three purposes: \textit{i)} it can extract similar visual semantics within one batch to optimize the input of our spiking model; \textit{ii)} more informative samples can be found through contrastive hard negative mining; \textit{iii)} it benefits learning ITM pre-training tasks, making it easier to understand and align the image and text semantic features. We calculate the similarity function to make parallel image-text pairs achieve higher similarity scores. Then, image and text queues are used to increase the number of negative examples inspired by MoCo \cite{ref33}.\\

\textit{2) Spiking to Text Uni-Alignment Learning (STUA):} The above ITC pre-training task enables the model to obtain similar image-text pairs in low-dimensional space. To this end, we leverage the input text semantics to assist images as candidates. Specifically, an alignment function is learned to project SNNs-generated visual features into text embeddings, defined as:
\begin{equation}
\label{eq:9}  
r(S,L)=h_{v}(\textbf{v}_{emb})^{T}h_{w}(\textbf{w}_{emb})
\end{equation}
where $h_v$ and $h_w$ represent linear transformations that map both of the embeddings to normalized lower-dimensional (256-$D$) representation. $\textbf{v}_{emb}=\{\textbf{v}_1, \dots, \textbf{v}_{N}\}$ represents the input image $S$ generated by the SNNs encoded into a sequence of embeddings. $\textbf{w}_{emb}=\{\textbf{w}_{1},\dots, \textbf{w}_{N}\}$ represents the sequence of embeddings transformed for the input text $L$.

For image $S$ and text $L$ embedding within one batch, the softmax-normalized spiking-to-text alignment score is calculated as:
\begin{equation}
\label{eq:10}  
f_{m}^{s2l}(S)=softmax(\frac{r(S,L_{m})}{\tau})
\end{equation}
where $r$ is a similarity function, and $\tau$ is a learnable temperature parameter. 

Let $p^{s2t}(S)$ represent the ground-truth one-hot similarity, the image text contrastive loss is expressed as the cross-entropy $H$:

\begin{equation}
\label{eq:11}  
\pounds_{STUA}=\frac{1}{2}\mathbb{E}_{{(S,L)}\backsim D}\lbrack H(p^{s2l}(S),(f^{s2l}(S))\rbrack
\end{equation}

\section{Experimental Setup}
\subsection{Implementation details}
We pre-train the model on 8 NVIDIA A100 GPUs with batch size 16 image-text pairs for 40 epochs. Our model is obtained through an underlying visual semantic encoder, which is used to initialize the parameters from ImageNet \cite{ref34}. To be specific, we use the ResNet-152 backbone and a 12-layer transformer architecture. For transformer, AdamW is used with learning rate $1\times 10^{-4}$ and weight decay $1\times 10^{-2}$. For the CNN backbone, SGD is selected as an optimizer with a learning rate of $1\times 10^{-2}$  and weight decay of $5\times 10^{-4}$. Furthermore, SGD is adopted with an initial learning rate of $1\times 10^{-2}$ and momentum of 0.9 as an SNNs optimizer, where the learning rate decays by 10 times at $25$th and $35$th training epochs. Besides, the time window $T$ is set to 10 (\textit{More details see ablation study}). For language features, the encoder is initialized using the WordPiece tokenizer \cite{ref35} based on BERT.

During pre-training tasks, we apply ITM, MLM, MVM, and ITC to achieve pre-processing, where the MVM and MLM are applied to the positive image-text pairs. For the first 10 epochs, we freeze part of the SNNs model parameters and choose to train the transformer, which means that concrete information is used for image features. After the first 10 epochs are completed, we perform ITC to learn similar visual semantics for processing batch images and training the hierarchical network model with STUA.

\subsection{Datasets and evaluation metric}
In our work, the pre-training datasets are built with VG and MSCOCO for many downstream VL tasks, which includes VQA, image-text retrieval, NLVR and visual entailment (VE). Table \ref{table_1} provides the usage of our pre-training datasets and different downstream tasks. 
\begin{table}[!t]
\centering
\renewcommand{\arraystretch}{1.3}
\setlength{\abovecaptionskip}{0.3cm}
\caption{Statistics of the different downstream tasks. Note that “$\blacktriangle$” denotes Karpathy split.}
\label{table_1}
\begin{IEEEeqnarraybox}[\IEEEeqnarraystrutmode\IEEEeqnarraystrutsizeadd{2pt}{2pt}]{l/c/c/c}
\IEEEeqnarrayrulerow[1pt]\\
\mbox{Task} ~& \mbox{Dataset} ~~& \mbox{Train Split} ~~&\mbox{Test Split} \\
\IEEEeqnarrayrulerow[0.5pt]\\
\raisebox{-7pt}[0pt][0pt]{Our pre-training}  ~~& \mbox{VG} ~& \mbox{train} ~~& \mbox{-}  \\
~& \mbox{MSCOCO} ~~&  \mbox{train+restval$^{\blacktriangle}$} ~~&\mbox{-} \\
\IEEEeqnarrayrulerow[0.5pt]\\
\raisebox{-12pt}[0pt][0pt]{VQA}  ~~& \mbox{VQA v2} ~& \mbox{train+val} ~~& \mbox{test-dev/test-std}  \\
~& \mbox{COCO-QA} ~~&  \mbox{train} ~~&\mbox{test-std} \\
~& \mbox{VQA-CP v2} ~~&  \mbox{train} ~~&\mbox{test-std} \\
\IEEEeqnarrayrulerow[0.5pt]\\
\raisebox{-7pt}[0pt][0pt]{Image-text Retrieval}  ~~& \mbox{MSCOCO} ~& \mbox{train+restval$^{\blacktriangle}$} ~~& \mbox{test$^{\blacktriangle}$}  \\
~& \mbox{Flickr30K} ~~&  \mbox{train} ~~&\mbox{test$^{\blacktriangle}$} \\
\IEEEeqnarrayrulerow[0.5pt]\\
\mbox{NLVR} ~& \mbox{NLVR$^2$} ~~& \mbox{train} ~~&\mbox{dev/test-P} \\
\IEEEeqnarrayrulerow[0.5pt]\\
\mbox{VE} ~& \mbox{SNLI-VE} ~~& \mbox{train} ~~&\mbox{val/test} \\
\IEEEeqnarrayrulerow[1pt]\\
\end{IEEEeqnarraybox}
\vspace*{-17pt}
\end{table}   
\textit{1) VQA:} To evaluate the specific task in VQA, we use three publicly available large VQA datasets, namely VQA v2 \cite{ref36}, COCO-QA \cite{ref37}, and VQA-CP v2 \cite{ref38}. Specifically, VQA v2, as the most widely used typical multimodal dataset, contains approximately 3000 answers. The answers are divided into three categories: “\textit{Yes/No}”, “\textit{Number}”, and “\textit{Other}”. Here, the dataset generates 10 ground-truth answers per image-question pair from human annotators. An accuracy-based evaluation metric is defined as follows: 
\begin{equation}
\label{eq:12}  
Accuracy(a)=min(\frac{\#humans\ that\ said\ (a)}{3},1)
\end{equation}
where $a$ is the number of answers provided by different annotators. If the predicted answer matches a ground truth answer, the accuracy will be $1/3$. If the predicted answer is given by more than three annotators, the corresponding score is “1”.

COCO-QA is another widely used VQA dataset, which contains four different types of questions such as “\textit{Object}”, “\textit{Number}”, “\textit{Color}”, and “\textit{Location}”. Furthermore, we use the VQA-CP v2 dataset, which is the well-known benchmark for changing the prior distributions of answers to overcome the question-oriented bias problem. 

\textit{2) Image-text Retrieval:} Image-text retrieval can be divided into two subtasks, image-to-text retrieval (TR) and text-to-image retrieval (IR). The proposed model is trained to evaluate the performance on the Flickr30K \cite{ref39} and MSCOCO datasets. Flickr30K consists of 31,783 images, where each image is described with 5 sentences. MSCOCO contains 123,287 images and each is annotated with 5 descriptions. We use Recall@K (K=1,5,10) as the evaluation metrics in the experiments. Specifically, R@K is the percentage of correct matchings in the top-K lists. The higher R@K indicates better performance. 
 
\textit{3) NLVR:} We conduct the NLVR task on the NLVR$^2$ datasets \cite{ref40}, which contains approximately 219.9k image-text pairs. NLVR$^2$ takes a pair of images and a natural language, and the goal is to predict whether a text statement about the image pair is correct. Compared with VQA, NLVR NLVR focuses more on reasoning relations and quantities. In our experiments, the settings of learning rate, epoch, and optimizer are the same as VQA settings.
 
\textit{4) VE:} The task of VE is performed on the SNLI-VE datasets \cite{ref41}, which are constructed based on Flickr30K and  Stanford Natural Language Inference (SNLI) datasets. The training, validation, and test sets contain 29.8k/1.0k/1.0k images, respectively.
\begin{table*}[!t]
\centering
\renewcommand{\arraystretch}{1.3}
\setlength{\abovecaptionskip}{0.3cm}
\caption{Comparison with the state-of-the-art approaches on the VQA v2.}
\label{table_2}
\begin{IEEEeqnarraybox}[\IEEEeqnarraystrutmode\IEEEeqnarraystrutsizeadd{1pt}{1pt}]{l/c/c/c/c/c/c/c/c/c/c/c/c}
\IEEEeqnarrayrulerow[1pt]\\
\raisebox{-9pt}[0pt][0pt]{Method}~~&\raisebox{-9pt}[0pt][0pt]{Transformer}~~~&\IEEEeqnarraymulticol{4}{t}{\mbox{Test-dev (\%)}}&&\IEEEeqnarraymulticol{5}{t}{\mbox{Test-std (\%)}}\\
\cmidrule[0.5pt]{3-6}\cmidrule[0.5pt]{8-13}
&&\mbox{Y/N}~~~~~~&\mbox{Num}~~~~~~&\mbox{Other}~~~~~~&\mbox{Overall} &~~~&\mbox{Y/N}~~~~~~&\mbox{Num}~~~~~~&\mbox{Other}~~~~~~&\mbox{Overall} \\  
\IEEEeqnarrayrulerow[0.5pt]\\
\mbox{UpDn \cite{ref6}}~~&\usym{1F5F4}~~~~&81.82~~~~~~&44.21~~~~~~&56.05~~~~~~&65.32&~~~&82.20~~~~~~&43.90~~~~~~&56.26~~~~~~&65.67\\                                                                            
\mbox{CRA-Net \cite{ref42}}~~&\usym{1F5F4}~~~~&84.87~~~~~~&49.46~~~~~~&59.08~~~~~~&68.61&~~~&85.21~~~~~~&48.43~~~~~~&59.42~~~~~~&68.92\\    
\mbox{ReGAT \cite{ref43}}~~&\usym{1F5F4}~~~~&86.08~~~~~~&54.42~~~~~~&60.33~~~~~~&70.27&~~~&-~~~~~~&-~~~~~~&-~~~~~~&-\\
\mbox{ViLBERT \cite{ref44}}~~&\usym{2713}~~~~&-~~~~~~&-~~~~~~&-~~~~~~&70.55&~~~&-~~~~~~&-~~~~~~&-~~~~~~&70.92\\
\mbox{PTP-ViLT \cite{ref45}}~~&\usym{2713}~~~~&-~~~~~~&-~~~~~~&-~~~~~~&72.13&~~~&-~~~~~~&-~~~~~~&-~~~~~~&73.36\\ 
\mbox{LXMERT \cite{ref46}}~~&\usym{2713}~~~~&87.00~~~~~~&52.66~~~~~~&61.57~~~~~~&72.42&~~~&88.20~~~~~~&54.20~~~~~~&63.10~~~~~~&72.50\\ 
\mbox{UNITER \cite{ref14}}~~&\usym{2713}~~~~&-~~~~~~&-~~~~~~&-~~~~~~&72.70&~~~&-~~~~~~&-~~~~~~&-~~~~~~&72.91\\ 
\mbox{TRAR \cite{ref47}}~~&\usym{2713}~~~~&88.11~~~~~~&55.33~~~~~~&63.31~~~~~~&72.93&~~~&-~~~~~~&-~~~~~~&-~~~~~~&-\\ 
\mbox{SOHO \cite{ref11}}~~&\usym{2713}~~~~&-~~~~~~&-~~~~~~&-~~~~~~&73.25&~~~&-~~~~~~&-~~~~~~&-~~~~~~&73.47\\ 
\IEEEeqnarrayrulerow[0.5pt]\\
\IEEEeqnarrayseprow[2pt]\\
\mbox{\textbf{ASH-Nets (Ours)}}~~&\usym{2713}~~~~&\textbf{91.92}~~~~~~&\textbf{58.33}~~~~~~&\textbf{65.45}~~~~~~&\textbf{75.40}&~~~&\textbf{92.00}~~~~~~&\textbf{59.10}~~~~~~&\textbf{68.98}~~~~~~&\textbf{75.57}\\
\IEEEeqnarrayrulerow[1pt]\\
\end{IEEEeqnarraybox}
\vspace*{-7pt}
\end{table*} 

\section{Experimental Results}
In this section, we evaluate the overall superiority of the developed model on three typical benchmark datasets. On this basis, we provide a detailed analysis to demonstrate the speed advantage on low-power devices, and the application prospects of our framework.
\begin{table*}[!t]
\centering
\renewcommand{\arraystretch}{1.3}
\setlength{\abovecaptionskip}{0.3cm}
\caption{Comparison with the state-of-the-art approaches on the COCO-QA.}
\label{table_3}
\begin{IEEEeqnarraybox}[\IEEEeqnarraystrutmode\IEEEeqnarraystrutsizeadd{2pt}{2pt}]{l/c/c/c/c/c/c}
\IEEEeqnarrayrulerow[1pt]\\
\mbox{Method}~~~~~&\mbox{Object}~~~~~~~~&\mbox{Number}~~~~~~~~&\mbox{Color} ~~~~~~~~&\mbox{Location}~~~~~~&\mbox{Other}~~~~~~&\mbox{Overall}\\       
\IEEEeqnarrayrulerow[0.5pt]\\    
\mbox{UpDn}\mbox{\cite{ref6}}~~~~~~~&70.48 ~~~~~~~~&54.70 ~~~~~~~~&74.17 ~~~~~~~~&60.90 ~~~~~~&85.05~~~~~~&69.33\\ 
\mbox{MRA-Net}\mbox{\cite{ref48}}~~~~~~~&71.40  ~~~~~~~~&56.42  ~~~~~~~~&74.69  ~~~~~~~~&60.62   ~~~~~~&86.83  ~~~~~~&70.27\\                         
\IEEEeqnarrayrulerow[0.5pt]\\
\mbox{\textbf{ASH-Nets (Ours)}} ~~~~~~~&\textbf{74.83} ~~~~~~~~&\textbf{56.92} ~~~~~~~~&\textbf{76.99}~~~~~~~~&\textbf{64.13}  ~~~~~~&\textbf{89.96}~~~~~~&\textbf{74.02} \\                      
\IEEEeqnarrayrulerow[1pt]\\
\end{IEEEeqnarraybox}
\vspace*{-10pt}
\end{table*}  

\begin{table}[!t]
\centering
\renewcommand{\arraystretch}{1.3}
\setlength{\abovecaptionskip}{0.3cm}
\caption{Comparison with the state-of-the-art approaches \\on the VQA-CP v2.}
\label{table_4}   
\begin{IEEEeqnarraybox}[\IEEEeqnarraystrutmode\IEEEeqnarraystrutsizeadd{2pt}{2pt}]{l/c/c/c/c/c}
\IEEEeqnarrayrulerow[1pt]\\ 
\mbox{Method} ~~~&\mbox{Y/N}~~~~&\mbox{Num} ~~~~&\mbox{Other}~~~~&\mbox{Overall}\\
\IEEEeqnarrayrulerow[0.5pt]\\
\mbox{ReGAT}\mbox{\cite{ref43}}~~~&-~~~~&-~~~~&-~~~~&40.42\\ 
\mbox{MRA-Net}\mbox{\cite{ref48}}~~~&44.53 ~~~~&13.05 ~~~~ &45.83 ~~~~ &40.45\\
\mbox{UpDn+LPF}\mbox{\cite{ref49}}~~~&88.62 ~~~~&23.78~~~~&46.57~~~~&55.34\\
\mbox{GGE-DQ}\mbox{\cite{ref50}}~~~&87.35~~~~&26.16~~~~&49.77~~~~&57.12\\
\IEEEeqnarrayrulerow[0.5pt]\\
\textbf{\mbox{ASH-Nets (Ours)}}~~~&\textbf{89.92} ~~~~&\textbf{38.44} ~~~~&\textbf{50.47} ~~~~&\textbf{58.99}\\
\IEEEeqnarrayrulerow[1pt]\\
\end{IEEEeqnarraybox}
\vspace*{-15pt}
\end{table} 

\subsection{Performance comparison on VQA task}
We compare the proposed model with the state-of-the-art VQA models, where UpDn \cite{ref6} is a baseline model, relying on the traditional region-based detection method. UNITER \cite{ref14} is a powerful pre-training model that uses a 24-layer transformer as a language module. SOHO develops a visual dictionary. As shown in Table \ref{table_2}, compared to the TRAR \cite{ref47} model on the test-dev set, our model improves the performance by 2.47\%. In addition, our model increases the overall accuracy of UNITER by 2.7\% on the test-dev set and by 2.66\% accuracy on the test-std set. Note that our result is even higher than the performance of SOHO. These obviously indicate the effectiveness of the proposed hierarchical network model to reduce signal spaces difference during the process of VL alignment. In Table \ref{table_3} and Table \ref{table_4}, we conduct additional experiments on COCO-QA and VQA-CP v2 datasets, where VQA-CP v2 is applied for overcoming the question-oriented bias. Obviously, our comparison results outperform other methods.

\subsection{Other downstream tasks and results}
In this section, we also evaluate the effectiveness of our model on other VL tasks, such as Image-text retrieval, NLVR, and VE. Table \ref{table_5} and Table \ref{table_6} report results on Image-text retrieval, which evaluate two subtasks TR and IR on MSCOCO and Flickr30K datasets, respectively. Compared with the state-of-the-art methods, we achieve better accuracy to fully validate the superiority of the proposed ASH-Nets when applied to Image-text retrieval tasks. Table \ref{table_7} shows the experimental results for visual reasoning task. Compared with the ALBEF \cite{ref30} and X-VLM \cite{ref51}, we achieve absolute improvements of 5.59\% and 1.88\% on NLVR$^2$ test-P. Furthermore, 5.71\% and 1.76\% absolute gains of ASH-Nets against ALBEF and X-VLM on dev. In Table \ref{table_8}, comparison results on the SNLI-VE dataset are displayed. As shown in Table \ref{table_8}, our proposed model achieves  89.51\% and 89.05\% accuracy on val and test respectively, outperforming  recent competing models. The promising results demonstrate that our end-to-end pre-training framework enables to facilitate VL alignment.

\begin{table*}[!t]
\centering
\renewcommand{\arraystretch}{1.3}
\setlength{\abovecaptionskip}{0.3cm}
\caption{Image-text retrieval comparison results with the state-of-the-art approaches on the MSCOCO dataset.}
\label{table_5}
\begin{IEEEeqnarraybox}[\IEEEeqnarraystrutmode\IEEEeqnarraystrutsizeadd{1pt}{1pt}]{l/c/c/c/c/c/c/c/c/c/c/c/c/c/c}
\IEEEeqnarrayrulerow[1pt]\\
\raisebox{-9pt}[0pt][0pt]{Method}~~&\raisebox{-9pt}[0pt][0pt]{Backbone}~~~&\IEEEeqnarraymulticol{3}{t}{\mbox{TR}}&\IEEEeqnarraymulticol{3}{t}{\mbox{IR}}&\IEEEeqnarraymulticol{3}{t}{\mbox{TR}}&\IEEEeqnarraymulticol{3}{t}{\mbox{IR}}\\
\cmidrule[0.5pt]{3-5}\cmidrule[0.5pt]{6-8}\cmidrule[0.5pt]{9-11}\cmidrule[0.5pt]{12-15}
&&\mbox{R@1}~~~~&\mbox{R@5}~~~~&\mbox{R@10}~~~~&\mbox{R@1}~~~~&\mbox{R@5}~~~~&\mbox{R@10}~~~~&\mbox{R@1}~~~~&\mbox{R@5}~~~~&\mbox{R@10}~~~~&\mbox{R@1}~~~~&\mbox{R@5}~~~~&\mbox{R@10} \\  
\IEEEeqnarrayrulerow[0.5pt]\\
&&\IEEEeqnarraymulticol{6}{t}{\mbox{1K Test set}}&\IEEEeqnarraymulticol{6}{t}{\mbox{5K Test set}}\\
\IEEEeqnarrayrulerow[0.5pt]\\
\mbox{SCAN \cite{ref52}}~~&\mbox{R101}~~~&72.70~~~~&94.80~~~~&98.40~~~~&58.80~~~~&88.40~~~~&94.80~~~~&50.40~~~~&82.20~~~~&90.00~~~~&38.60~~~~&69.30~~~~&80.40\\ 
\mbox{Unicoder-VL \cite{ref53}}~~&-~~~&84.30~~~~&97.30~~~~&99.30~~~~&69.70~~~~&93.50~~~~&97.20~~~~&62.30~~~~&87.10~~~~&92.80~~~~&46.70~~~~&76.00~~~~&85.30\\ 
\mbox{UNITER \cite{ref14}}~~&\mbox{R101}~~~&-~~~~&-~~~~&-~~~~&-~~~~&-~~~~&-~~~~&64.40~~~~&87.40~~~~&93.10~~~~&50.30~~~~&78.50~~~~&87.20 \\ 
\mbox{SOHO \cite{ref11}}~~&\mbox{R101}~~~&85.10~~~~&97.40~~~~&99.40~~~~&73.50~~~~&94.50~~~~&97.50~~~~&66.40~~~~&88.20~~~~&93.80~~~~&50.60~~~~&78.00~~~~&86.70 \\ 
\mbox{VISTA \cite{ref54}}~~&\mbox{ViT-B/16}~~~&-~~~~&-~~~~&-~~~~&-~~~~&-~~~~&-~~~~&68.90~~~~&90.10~~~~&95.40~~~~&52.60~~~~&79.60~~~~&87.60 \\ 
\mbox{OmniVL \cite{ref55}}~~&\mbox{ViT-B/16}~~~&-~~~~&-~~~~&-~~~~&-~~~~&-~~~~&-~~~~&82.10~~~~&95.90~~~~&98.10~~~~&64.80~~~~&86.10~~~~&91.60 \\ 
\IEEEeqnarrayrulerow[0.5pt]\\
\mbox{\textbf{ASH-Nets (Ours)}}~~&\mbox{R152}~~~&\textbf{90.30}~~~~&\textbf{98.20}~~~~&\textbf{99.60}~~~~&\textbf{75.90}~~~~&\textbf{95.90}~~~~&\textbf{98.00}~~~~&\textbf{82.90}~~~~&\textbf{96.20}~~~~&\textbf{98.20}~~~~&\textbf{65.30}~~~~&\textbf{87.00}~~~~&\textbf{92.20} \\ 
\IEEEeqnarrayrulerow[1pt]\\
\end{IEEEeqnarraybox}
\vspace*{-7pt}
\end{table*} 

\begin{table*}[!t]
\centering
\renewcommand{\arraystretch}{1.3}
\setlength{\abovecaptionskip}{0.3cm}
\caption{Image-text retrieval comparison results with the state-of-the-art approaches \\on the Flickr30K dataset.}
\label{table_6}
\begin{IEEEeqnarraybox}[\IEEEeqnarraystrutmode\IEEEeqnarraystrutsizeadd{1pt}{1pt}]{l/c/c/c/c/c/c/c/c}
\IEEEeqnarrayrulerow[1pt]\\
\raisebox{-9pt}[0pt][0pt]{Method}~~~&\raisebox{-9pt}[0pt][0pt]{Backbone}~~~~~~&\IEEEeqnarraymulticol{3}{t}{\mbox{TR}}&~~~&\IEEEeqnarraymulticol{3}{t}{\mbox{IR}}\\
\cmidrule[0.5pt]{3-5}\cmidrule[0.5pt]{7-9}
&&\mbox{R@1}~~~~&\mbox{R@5}~~~~&\mbox{R@10}~~~~&~~~~&\mbox{R@1}~~~~&\mbox{R@5}~~~~&\mbox{R@10} \\  
\IEEEeqnarrayrulerow[0.5pt]\\
\mbox{SCAN \cite{ref52}}~~~&\mbox{R101}~~~~&67.40~~~~&90.30~~~~&95.80~~~~&~~~~&48.60~~~~&77.70~~~~&85.20\\ 
\mbox{Unicoder-VL \cite{ref53}}~~~&-~~~~&86.20~~~~&96.30~~~~&99.00~~~~&~~~~&71.50~~~~&90.90~~~~&94.90\\ 
\mbox{ViLBERT \cite{ref44}}~~~&\mbox{R101}~~~~&-~~~~&-~~~~&-~~~~&~~~~&58.20~~~~&84.90~~~~&91.50\\ 
\mbox{UNITER \cite{ref14}}~~~&\mbox{R101}~~~~&85.90~~~~&97.10~~~~&98.80~~~~&~~~~&72.50~~~~&92.40~~~~&96.10\\ 
\mbox{SOHO \cite{ref11}}~~~&\mbox{R101}~~~~&86.50~~~~&98.10~~~~&99.30~~~~&~~~~&72.50~~~~&92.70~~~~&96.10\\ 
\mbox{VISTA \cite{ref54}}~~~&\mbox{ViT-B/16}~~~~&89.50~~~~&98.40~~~~&99.60~~~~&~~~~&75.80~~~~&94.20~~~~&96.90\\ 
\mbox{NAPReg \cite{ref56}}~~~&-~~~~&60.00~~~~&84.10~~~~&90.20~~~~&~~~~&79.60~~~~&95.60~~~~&98.00\\ 
\IEEEeqnarrayrulerow[0.5pt]\\
\mbox{\textbf{ASH-Nets (Ours)}}~~~&\mbox{R152}~~~~&\textbf{90.10}~~~~&\textbf{98.50}~~~~&\textbf{99.60}~~~~&~~~~&\textbf{79.60}~~~~&\textbf{94.90}~~~~&\textbf{98.20} \\ 
\IEEEeqnarrayrulerow[1pt]\\
\end{IEEEeqnarraybox}
\vspace*{-7pt}
\end{table*} 

\begin{table}[!t]
\centering
\renewcommand{\arraystretch}{1.3}
\setlength{\abovecaptionskip}{0.3cm}
\caption{NLVR comparison results with the state-of-the-art \\approaches on the NLVR$^2$ dataset.}
\label{table_7}   
\begin{IEEEeqnarraybox}[\IEEEeqnarraystrutmode\IEEEeqnarraystrutsizeadd{2pt}{2pt}]{l/c/c/c}
\IEEEeqnarrayrulerow[1pt]\\ 
\mbox{Method} ~~~~~&\mbox{Backbone}~~~~~&\mbox{dev} ~~~~~&\mbox{test-P}\\
\IEEEeqnarrayrulerow[0.5pt]\\
\mbox{MaxEnt}\mbox{\cite{ref40}}~~~~~&\mbox{R152}~~~~~&54.10~~~~~&54.80\\ 
\mbox{VisualBERT}\mbox{\cite{ref57}}~~~~~&\mbox{R152}~~~~~&67.40~~~~~&67.00\\ 
\mbox{LXMERT}\mbox{\cite{ref46}}~~~~~&\mbox{R101} ~~~~~&74.90~~~~~&74.50\\
\mbox{UNITER}\mbox{\cite{ref14}}~~~~~&\mbox{R101} ~~~~~&75.85 ~~~~~ &75.80 \\
\mbox{SOHO}\mbox{\cite{ref11}}~~~~~&\mbox{R101}~~~~~&76.37~~~~~&77.32\\
\mbox{SimVLM}\mbox{\cite{ref58}}~~~~~&\mbox{ViT-B/16}~~~~~&81.72~~~~~&81.77\\
\mbox{ALBEF}\mbox{\cite{ref30}}~~~~~&\mbox{ViT-B/16}~~~~~&80.21~~~~~&80.50\\
\mbox{X-VLM}\mbox{\cite{ref51}}~~~~~&\mbox{ViT-B/16}~~~~~&84.16~~~~~&84.21\\
\IEEEeqnarrayrulerow[0.5pt]\\
\mbox{\textbf{ASH-Nets (Ours)}}~~~~~&\mbox{R152}~~~~~&\textbf{85.92}~~~~~&\textbf{86.09}\\
\IEEEeqnarrayrulerow[1pt]\\
\end{IEEEeqnarraybox}
\vspace*{-1pt}
\end{table}
\begin{table}[!t]
\centering
\renewcommand{\arraystretch}{1.3}
\setlength{\abovecaptionskip}{0.3cm}
\caption{VE comparison results with the state-of-the-art \\approaches on the SNLI-VE dataset.}
\label{table_8}   
\begin{IEEEeqnarraybox}[\IEEEeqnarraystrutmode\IEEEeqnarraystrutsizeadd{2pt}{2pt}]{l/c/c/c}
\IEEEeqnarrayrulerow[1pt]\\ 
\mbox{Method} ~~~~~&\mbox{Backbone}~~~~~&\mbox{val} ~~~~~&\mbox{test}\\
\IEEEeqnarrayrulerow[0.5pt]\\
\mbox{UNITER}\mbox{\cite{ref14}}~~~~~&\mbox{R101} ~~~~~&78.59 ~~~~~ &78.28 \\
\mbox{SOHO}\mbox{\cite{ref11}}~~~~~&\mbox{R101}~~~~~&85.00~~~~~&84.95\\
\mbox{Prompt Tuning}\mbox{\cite{ref59}}~~~~~&-~~~~~&88.59~~~~~&88.18\\
\mbox{CoCa}\mbox{\cite{ref60}}~~~~~&\mbox{ViT-B/16}~~~~~&87.00~~~~~&87.10\\
\IEEEeqnarrayrulerow[0.5pt]\\
\mbox{\textbf{ASH-Nets (Ours)}}~~~~~&\mbox{R152}~~~~~&\textbf{89.51}~~~~~&\textbf{89.05}\\
\IEEEeqnarrayrulerow[1pt]\\
\end{IEEEeqnarraybox}
\vspace*{-7pt}
\end{table} 
\begin{table}[!t]
\centering
\renewcommand{\arraystretch}{1.3}
\setlength{\abovecaptionskip}{0.3cm}
\caption{Evaluation of downstream VL tasks at different retrieve numbers of similar samples.}
\label{table_9}
\begin{IEEEeqnarraybox}[\IEEEeqnarraystrutmode\IEEEeqnarraystrutsizeadd{1pt}{1pt}]{l/c/c/c/c/c/c}
\IEEEeqnarrayrulerow[1pt]\\
\raisebox{-8pt}[0pt][0pt]{Numbers}~~~&\IEEEeqnarraymulticol{2}{t}{\raisebox{-1pt}[0pt][0pt]{MSCOCO 1K Test}}~&~~~&\raisebox{-1pt}[0pt][0pt]{VQA v2}~~~~~&\raisebox{-1pt}[0pt][0pt]{NLVR$^2$}~~~&\raisebox{-1pt}[0pt][0pt]{SNLI-VE}\\
\cmidrule[0.5pt]{2-8}
&\mbox{TR}~~~~~~~&\mbox{IR}~&~~~&\mbox{test-std}~~~~~&\mbox{dev}~~~~&\mbox{test}\\ 
\IEEEeqnarrayrulerow[0.5pt]\\
\mbox{4}~~~&88.50~~~~~~~&74.70~&~~~&63.01~~~~~&83.62~~~~&88.25\\                                                                             
\mbox{8}~~~&89.20~~~~~~~&75.00~&~~~&72.56~~~~~&84.92~~~~&88.93\\
\mbox{16}~~~&\textbf{90.10}~~~~~~~&\textbf{75.50}~&~~~&\textbf{74.28}~~~~~&\textbf{85.01}~~~~&\textbf{89.11}\\   
\mbox{32}~~~&90.30~~~~~~~&75.60~&~~~&70.67~~~~~&84.04~~~~&89.09\\
\mbox{64}~~~&90.30~~~~~~~&75.70~&~~~&70.19~~~~~&84.01~~~~&88.25\\ 
\IEEEeqnarrayrulerow[1pt]\\
\end{IEEEeqnarraybox}
\vspace*{-7pt}
\end{table}  
\begin{table*}[!t]
\centering
\renewcommand{\arraystretch}{1.3}
\setlength{\abovecaptionskip}{0.3cm}
\caption{Evaluation results for different time window sizes on VQA v2 dataset.}
\label{table_10}
\begin{IEEEeqnarraybox}[\IEEEeqnarraystrutmode\IEEEeqnarraystrutsizeadd{1pt}{1pt}]{l/c/c/c/c/c/c/c/c/c/c/c}
\IEEEeqnarrayrulerow[1pt]\\
\raisebox{-9pt}[0pt][0pt]{Window sizes}~~~~~&\IEEEeqnarraymulticol{4}{t}{\mbox{Test-dev (\%)}}&&\IEEEeqnarraymulticol{5}{t}{\mbox{Test-std (\%)}}\\
\cmidrule[0.5pt]{2-5}\cmidrule[0.5pt]{7-12}
&\mbox{Y/N}~~~~~~&\mbox{Num}~~~~~~&\mbox{Other}~~~~~~&\mbox{Overall} &~~~~~&\mbox{Y/N}~~~~~~&\mbox{Num}~~~~~~&\mbox{Other}~~~~~~&\mbox{Overall} \\  
\IEEEeqnarrayrulerow[0.5pt]\\
\mbox{1}~~~&80.78~~~~~~&52.44~~~~~~&51.22~~~~~~&63.91&~~~&81.05~~~~~~&46.74~~~~~~&54.12~~~~~~&62.92\\                                                                            
\mbox{5}~~~&89.12~~~~~~&55.28~~~~~~&62.35~~~~~~&72.28&~~~&88.57~~~~~~&56.13~~~~~~&65.98~~~~~~&71.99\\    
\mbox{10}~~~&\textbf{90.80}~~~~~~&\textbf{57.21}~~~~~~&\textbf{64.33}~~~~~~&\textbf{74.28}&~~~&\textbf{90.88}~~~~~~&\textbf{57.98}~~~~~~&\textbf{67.86}~~~~~~&\textbf{74.45}\\
\mbox{20}~~~&89.98~~~~~~&56.36~~~~~~&63.52~~~~~~&73.42&~~~&89.99~~~~~~&57.11~~~~~~&66.99~~~~~~&73.57\\
\IEEEeqnarrayrulerow[1pt]\\
\end{IEEEeqnarraybox}
\vspace*{-7pt}
\end{table*}  
\begin{table*}[!t]
\centering
\renewcommand{\arraystretch}{1.3}
\setlength{\abovecaptionskip}{0.3cm}
\caption{Performance comparison of four different hierarchical models.}
\label{table_11}
\begin{IEEEeqnarraybox}[\IEEEeqnarraystrutmode\IEEEeqnarraystrutsizeadd{1pt}{1pt}]{l/c/c/c/c/c/c/c}
\IEEEeqnarrayrulerow[1pt]\\
\raisebox{-8pt}[0pt][0pt]{Models}~~~~&\raisebox{-8pt}[0pt][0pt]{Parameter $SR$}~~~~&\IEEEeqnarraymulticol{2}{t}{\raisebox{-1pt}[0pt][0pt]{MSCOCO 1K Test}}~&~~~&\raisebox{-1pt}[0pt][0pt]{VQA v2}~~~~~~&\raisebox{-1pt}[0pt][0pt]{NLVR$^2$}~~~~~&\raisebox{-1pt}[0pt][0pt]{SNLI-VE}\\
\cmidrule[0.5pt]{3-9}
&&\mbox{TR}~~~~~~~&\mbox{IR}~&~~~&\mbox{test-std}~~~~~~&\mbox{dev}~~~~~~&\mbox{test}\\ 
\IEEEeqnarrayrulerow[0.5pt]\\
\mbox{ANNs}~~~~&\mbox{0}~~~~&85.20~~~~~~~&64.70~&~~~&62.92~~~~~~&80.59~~~~~~&84.00\\                                                                             
\mbox{ANNs+SNNs}~~~~&\mbox{1}~~~~&86.30~~~~~~~&65.90~&~~~&71.99~~~~~~&82.95~~~~~~&85.88\\
\mbox{SNNs}~~~~&-~~~~&60.70~~~~~~~&52.90~&~~~&56.33~~~~~~&70.99~~~~~~&72.17\\   
\mbox{ASH-Nets}~~~~&\mbox{trainable}~~~~&\textbf{90.10}~~~~~~~&\textbf{75.50}~&~~~&\textbf{74.45}~~~~~~&\textbf{85.01}~~~~~~&\textbf{89.11}\\
\IEEEeqnarrayrulerow[1pt]\\
\end{IEEEeqnarraybox}
\vspace*{-7pt}
\end{table*}
\begin{figure}[!t]
\begin{center}
\includegraphics[width=1.72in]{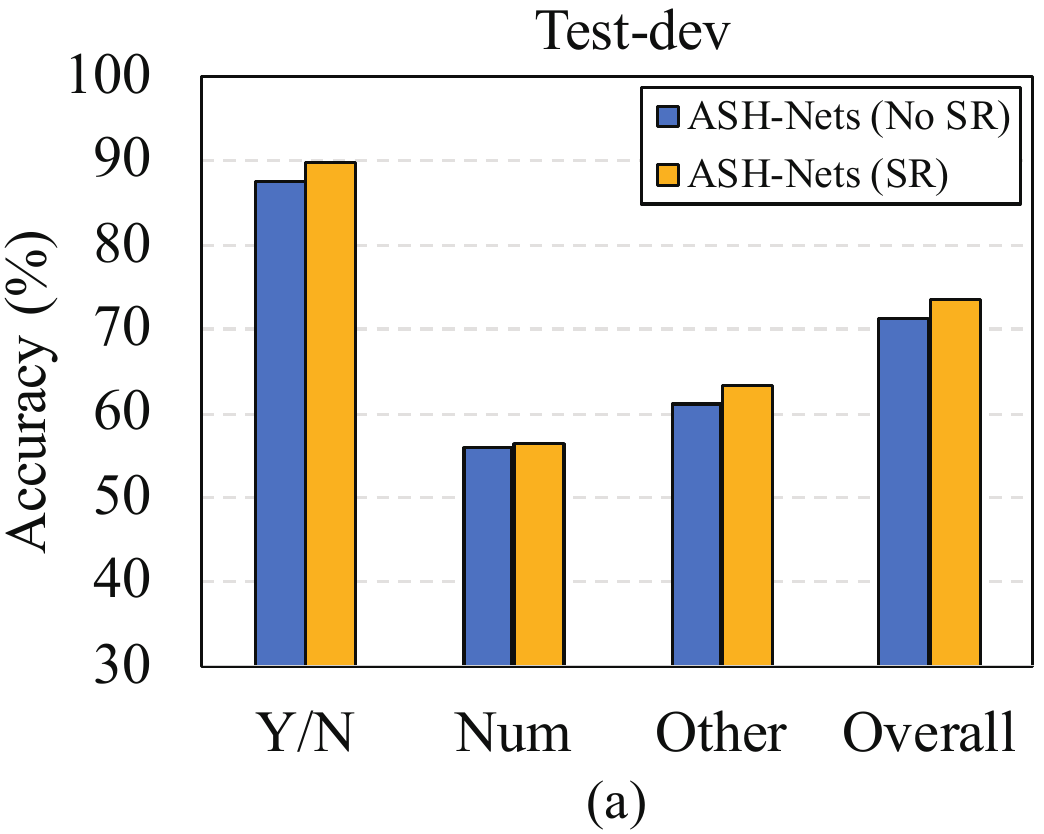}
\includegraphics[width=1.72in]{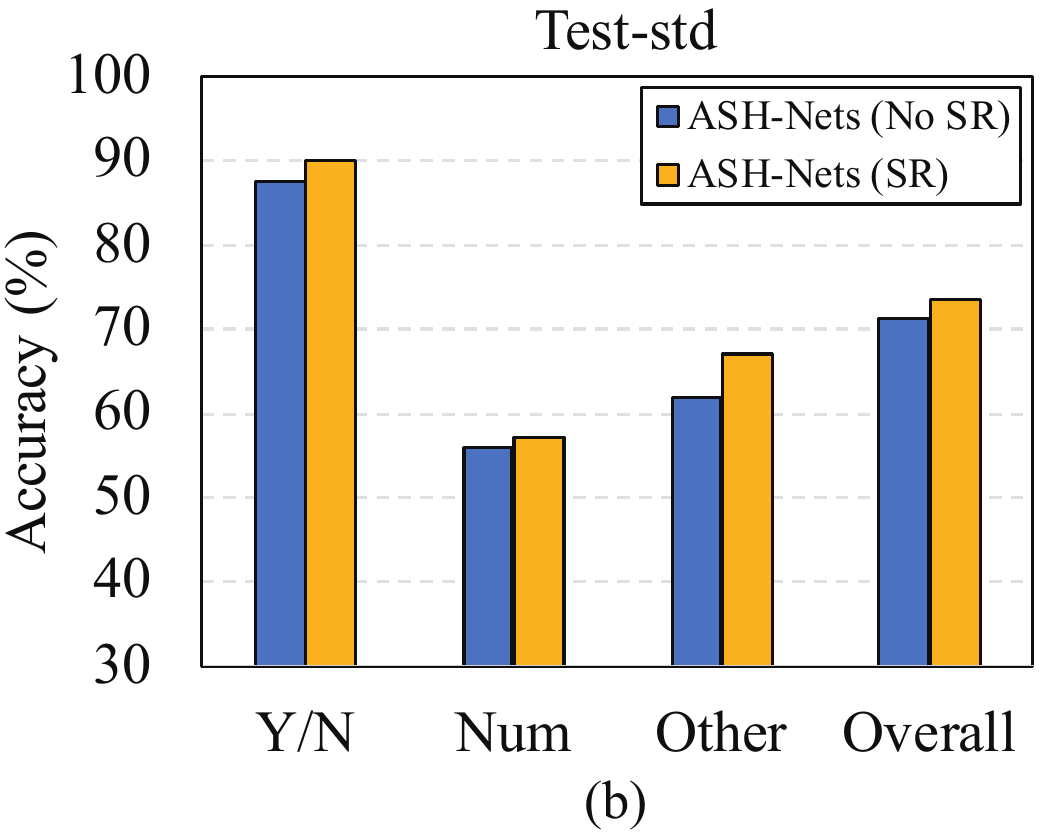}
\vspace*{-15pt}
\end{center}
   \caption{The accuracy of the proposed ASH-Nets model with and without the Summary ratio (SR).}
\label{fig_3}
\vspace*{-6pt}
\end{figure}
\begin{figure*}[t!]
\begin{center}
\includegraphics[width=2.3in]{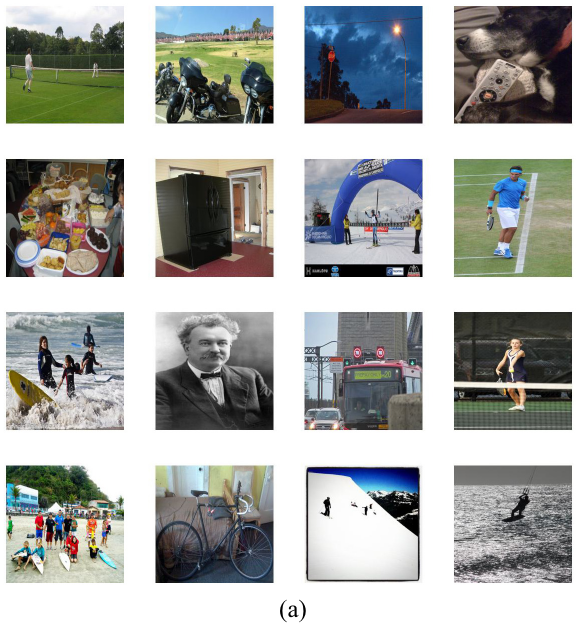}~~~~~~~~
\includegraphics[width=2.3in]{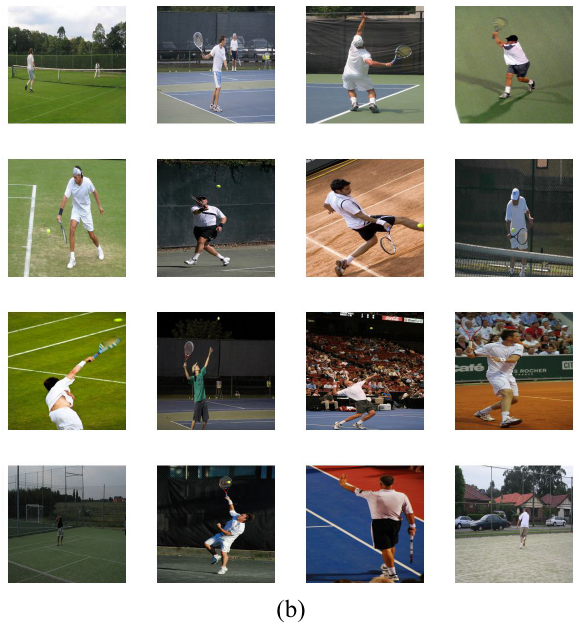}
\vspace*{-6pt}
\end{center}
   \caption{Comparison of the original and similar samples within a batch.}
\label{fig_4}
\end{figure*}
\subsection{Ablation study}
We further perform ablation studies to validate the effectiveness of each part in our proposed model. Here, a ResNet-101 backbone and 3-layer transformer are built to reduce computational cost. Note that all the evaluated models are trained on the VQA v2 dataset. First, the superiority with and without the Summary ratio (SR) is compared in our model. As shown in Fig. \ref{fig_3}, we found that the accuracy of the model with $SR$ improved. While for the model without $SR$, the inputs of the abstract semantic encoder within one batch are the same, which affects the effective processing of the “\textit{Num}” question. Also, this indirectly indicates that the abstract semantic encoder more focuses on the semantics itself. Additionally, the model with SR can achieve intra-modal feature fusion with fewer parameters and are more suitable for “\textit{Yes/No}” or “\textit{Other}” questions.

\begin{figure}[t!]
\begin{center}
\includegraphics[width=1.72in]{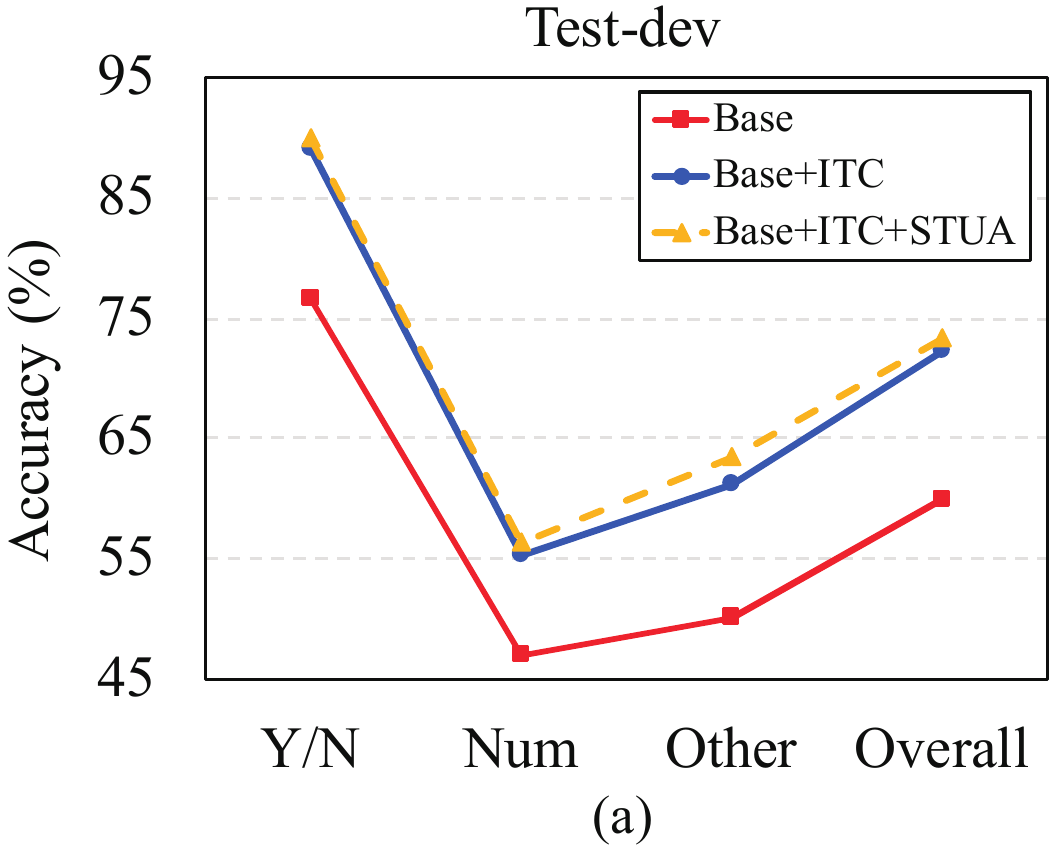}
\includegraphics[width=1.72in]{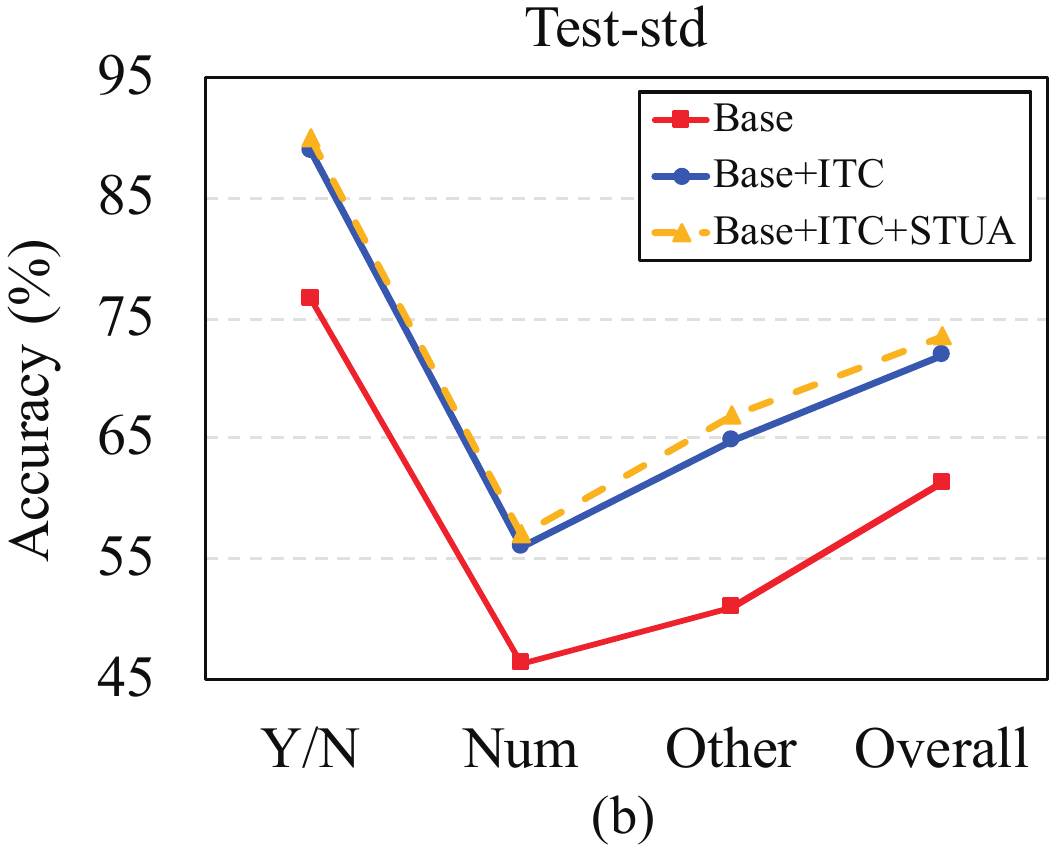}
\vspace*{-15pt}
\end{center}
   \caption{Performance comparison of different pre-training tasks, including “\textit{Base}”, “\textit{Base+ITC}”, and “\textit{Base+ITC+STUA}” three methods.}
\label{fig_5}
\end{figure}

In Table \ref{table_9}, we provide an experiment on the various downstream VL tasks, validating the performance of our model at different retrieve numbers of similar samples. Selecting the optimal hyperparameters is a key element in our model. For specific Image-text retrieval, higher retrieve numbers can learn more contrastive hard samples, which is beneficial for the model to extract deep features from similar samples. NLVR and VE have the same performance as the VQA task, and the following three aspects are considered: \textit{i)} higher retrieve numbers will over-recall similar images, which may include noisy semantics to reduce the similarity between images. \textit{ii)} higher retrieve numbers will overload the memory of our model. \textit{iii)} lower retrieve numbers produce a small number of similar semantic images, which will cause a waste of training efficiency by event-driven SNNs, and it is difficult to learn the commonality of abstract semantics. Hence, $Retrieve\ number=16$ is the best selection in our experiment. Fig. \ref{fig_4} (a) and (b) show the original samples within a batch and similar samples processed by using the ITC method, respectively. We perform an ablation study to analyze the effect of the pre-training tasks, as shown in Fig. \ref{fig_5}. Specifically, we provide three cases, including “\textit{Base}”, “\textit{Base+ITC}”, and “\textit{Base+ITC+STUA}”. The “\textit{Base}” model does not pre-process the samples in the batch, resulting in disordered activation output of the event-oriented (specific abstract semantic) SNNs model. Therefore, it is difficult to effectively learn specific semantics. The “\textit{Base+ITC}” model is more competitive than “\textit{Base}”. By contrast, the “\textit{Base+ITC+STUA}” model can further eliminate the semantic gap between image and language modalities, improving the performance of the model. 
\begin{table*}[!t]
\centering
\renewcommand{\arraystretch}{1.3}
\setlength{\abovecaptionskip}{0.3cm}
\caption{Comparison results of training times with different models. Note that “o” denotes the model\\without ITC. And “w” denotes the model with ITC.}
\label{table_12}
\begin{IEEEeqnarraybox}[\IEEEeqnarraystrutmode\IEEEeqnarraystrutsizeadd{1pt}{1pt}]{l/c/c/c/c/c/c/c}
\IEEEeqnarrayrulerow[1pt]\\
\raisebox{-1pt}[0pt][0pt]{Models}~~&\raisebox{-1pt}[0pt][0pt]{Training Epochs}~~~~&\raisebox{-1pt}[0pt][0pt]{Epoch Times (h)}~~~~&\raisebox{-1pt}[0pt][0pt]{Memory (mb)}~~~~&\raisebox{-1pt}[0pt][0pt]{Loss-MIM}~~~~&\raisebox{-1pt}[0pt][0pt]{Loss-MVM}~~~~&\raisebox{-1pt}[0pt][0pt]{Loss-Match}\\
\IEEEeqnarrayrulerow[0.5pt]\\
\mbox{ANNs}~~&\mbox{40}~~~~&2.3~~~~&19957~~~~&1.28~~~~&5.00~~~~&0.15\\                                                                             
\mbox{ASH-Nets (o)}~~&\mbox{100}~~~~&2.8~~~~&24023~~~~&1.31~~~~&5.10~~~~&0.16\\
\mbox{ASH-Nets (w)}~~&\mbox{30}~~~~&2.8~~~~&24023~~~~&1.30~~~~&4.98~~~~&0.15\\
\IEEEeqnarrayrulerow[1pt]\\
\end{IEEEeqnarraybox}
\vspace*{-7pt}
\end{table*}
\begin{table*}[!t]
\centering
\renewcommand{\arraystretch}{1.3}
\setlength{\abovecaptionskip}{0.3cm}
\caption{Performance comparison of different backbones on downstream VL tasks.}
\label{table_13}
\begin{IEEEeqnarraybox}[\IEEEeqnarraystrutmode\IEEEeqnarraystrutsizeadd{1pt}{1pt}]{l/c/c/c/c/c/c/c/c}
\IEEEeqnarrayrulerow[1pt]\\
\raisebox{-8pt}[0pt][0pt]{Backbone}~~~~&\raisebox{-8pt}[0pt][0pt]{Parameters (M)}~~~~&\raisebox{-8pt}[0pt][0pt]{Pre-training Times (h)}~~~~&\IEEEeqnarraymulticol{2}{t}{\raisebox{-1pt}[0pt][0pt]{MSCOCO 1K Test}}~&~~~&\raisebox{-1pt}[0pt][0pt]{VQA v2}~~~~~~&\raisebox{-1pt}[0pt][0pt]{NLVR$^2$}~~~~~&\raisebox{-1pt}[0pt][0pt]{SNLI-VE}\\
\cmidrule[0.5pt]{4-9}
&&&\mbox{TR}~~~~~~~&\mbox{IR}~&~~~&\mbox{test-dev}~~~~~~&\mbox{dev}~~~~~~&\mbox{test}\\ 
\IEEEeqnarrayrulerow[0.5pt]\\
\mbox{ResNet18}~~~~&11.69~~~~&72.50~~~~~~~&89.50~~~~&75.00~&~~~&74.01~~~~~~&84.59~~~~~~&88.28\\                                                                             
\mbox{ResNet101}~~~~&44.55~~~~&85.40~~~~~~~&90.10~~~~&75.50~&~~~&74.28~~~~~~&85.01~~~~~~&89.11\\
\mbox{ResNet152}~~~~&60.19~~~~&92.90~~~~~~~&90.30~~~~&75.90~&~~~&75.40~~~~~~&85.92~~~~~~&89.51\\   
\mbox{ViT-B/16}~~~~&391.12~~~~&202.10~~~~~~~&91.60~~~~&76.20~&~~~&77.52~~~~~~&86.95~~~~~~&89.93\\
\IEEEeqnarrayrulerow[1pt]\\
\end{IEEEeqnarraybox}
\vspace*{-7pt}
\end{table*}

In Table \ref{table_10}, to evaluate the impact of the time window sizes on the model, the window sizes are respectively set to 1, 5, 10, and 20. Given that we have designed the model with a special workload to improve the current SNNs signal transmission defects. It can be seen from Table \ref{table_10} that there is little difference when the time window size is set to 10 and 20. Notably, the proposed SNNs model only needs to analyze and extract the features in the semantic collector. Due to computational cost constraints, we set $Window\ size=10$ in this work. Table \ref{table_11} presents the performance comparison results of four different hierarchical models on various VL datasets, which contains “\textit{ANNs}”, “\textit{ANNs+SNNs}”, “\textit{SNNs}”, and “\textit{ASH-Nets}” architectures. We can observe that the accuracy of the model “\textit{SNNs}” is lower than other models. Since “\textit{SNNs}” only require to select the abstract semantic information of batch activations from the semantic collector during the pre-training process, it is difficult to implement other complex VL tasks.  “\textit{ASH-Nets}” (i.e., trainable $SR$) features achieve the best performance compared to fixed fusion “\textit{ANNs+SNNs}” and “\textit{ANNs}” models. Although the hyperparameter is set to 16, which can learn the maximum closeness of image semantics. However, it is not suitable to use the same pulse signal as an encoding of abstract semantics. To this end, we employ $SR$ to redistribute all signals, and the obtained different weights can make similar semantic images with different impulse signal representations. It is worth noting that the model “\textit{ANNs+SNNs}” is superior to the “\textit{ANNs}”, which indirectly proves the necessity and effectiveness of the SNNs structure.

Table \ref{table_12} compares the training time of three different models. The results of Table \ref{table_12} demonstrate the importance of similar samples calculated by ITC, which greatly reduces the epoch required for training. In addition, after $1/3$ of the training time and the same memory, a satisfactory loss can be achieved earlier by ASH-Nets. Table \ref{table_13} tests the impact of different backbone on the ASH-Nets model. We can see that the model accuracy improves as the number of parameters increases. Since the computational cost of time should not be ignored, we choose ResNet152 as our backbone. Furthermore, an experiment is designed in Table \ref{table_14} to verify the trainable semantic collector, which can enrich the output of the SNNs representations. More importantly, we consider that there are large differences in the feature spaces of SNNs and ANNs. However, the designed semantic collector and ANNs perform gradient descent by setting the same random initial value, that is. the vector space distribution is homologous. Therefore, the feature output of SNNs can be transformed into a combination of basis vectors for the semantic collector. This also fully reflects the spatial feature mapping of SNNs to the semantic collector, thereby fusing ANNs to achieve the final outputs.

\begin{table}[!t]
\centering
\renewcommand{\arraystretch}{1.3}
\setlength{\abovecaptionskip}{0.3cm}
\caption{Performance Evaluation of Semantic Collector. Note that “o” denotes the model without semantic collector. And “w” denotes the model with semantic collector.}
\label{table_14}
\begin{IEEEeqnarraybox}[\IEEEeqnarraystrutmode\IEEEeqnarraystrutsizeadd{1pt}{1pt}]{l/c/c/c/c/c/c}
\IEEEeqnarrayrulerow[1pt]\\
\raisebox{-4pt}[0pt][0pt]{Collector}~~~&\IEEEeqnarraymulticol{2}{t}{\raisebox{-1pt}[0pt][0pt]{MSCOCO 1K Test}}~~~~&\raisebox{-1pt}[0pt][0pt]{VQA v2}~~~~&\raisebox{-1pt}[0pt][0pt]{NLVR$^2$}~~~&\raisebox{-1pt}[0pt][0pt]{SNLI-VE}\\
\cmidrule[0.5pt]{2-8}
\raisebox{1pt}[0pt][0pt]{(w/o)}&\mbox{TR}~~~~~~~&\mbox{IR}~~~~&\mbox{test-dev}~~~~&\mbox{dev}~~~~&\mbox{test}\\ 
\IEEEeqnarrayrulerow[0.5pt]\\
\mbox{o}~~~&79.90~~~~~~~&63.40~~~~&63.28~~~~&76.25~~~~&80.07\\                                                                             
\mbox{w}~~~&90.10~~~~~~~&75.50~~~~&74.28~~~~&85.01~~~~&89.11\\
\IEEEeqnarrayrulerow[1pt]\\
\end{IEEEeqnarraybox}
\vspace*{-12pt}
\end{table}

\section{Conclusion}
In this work, we present an efficient ASH-Nets, which integrates visual concrete encoder and semantic abstract encoder to align vision-language tasks. More importantly, the proposed ASH-Nets successfully optimize visual embedding manner while enriching semantic representants. Next, we construct multi-layer transformers to fuse semantic features from VL modalities, which are applied to the VQA task to predict the corresponding answers. To effectively evaluate the performance of our model, we compare it with the state-of-the-art strategies by conducting extensive experiments on three benchmarks VQA datasets. The experimental results show that the designed method can better address the information asymmetry issue and achieve higher computational efficiency in multi-modal alignment. Moreover, our model is easily applied to VQA task outputs. In the future, we will further study different self-supervised based pre-training methods for VL alignment tasks.
 

\vfill

\begin{thebibliography}{60}
\bibliographystyle{IEEEtran}

\bibitem{ref1}
Y. Bi, H. Jiang, Y. Hu, Y. Sun, and B. Yin, “See and learn more: dense caption-aware representation for visual question answering”, \emph{IEEE Trans. Circuits Syst. Video Technol.}, Jul. 2023.
\bibitem{ref2}
S. Yang, Q. Li, W. Li, X. Li, and A. A. Liu, “Dual-Level Representation Enhancement on Characteristic and Context for Image-Text Retrieval”, \emph{IEEE Trans. Circuits Syst. Video Technol.}, vol. 32, no. 11, pp. 8037–8050, Nov. 2022.
\bibitem{ref3}
A. Suhr, M. Lewis, J. Yeh, and Y. Artzi, “A corpus of natural language for visual reasoning”, in \emph{Proc. 55th Annual Meeting of the ACL}, vol. 2, pp. 217-223, 2017.
\bibitem{ref4}
K. He, X. Zhang, S. Ren, and J. Sun, “Deep residual learning for image recognition”, in \emph{Proc. IEEE Conf. Comput. Vis. Pattern Recognit.}, pp. 770–778, 2016.
\bibitem{ref5}
K. Cho, B. V. Merrienboer, D. Bahdanau, and Y. Bengio, “On the properties of neural machine translation: encoder-decoder approaches”, 2014, \emph{arXiv:1409.1259}. [Online]. Available: http://arxiv.org/abs/1409.1259
\bibitem{ref6}
A. Peter \emph{et al.}, “Bottom-up and top-down attention for image captioning and visual question answering,” in \emph{Proc. IEEE Conf. Comput. Vis. Pattern Recognit.}, pp. 6077–6086, 2018.
\bibitem{ref7}
S. Ren, K. He, G. Ross, and J. Sun, “Faster R-CNN: towards real-time object detection with region proposal networks”, \emph{IEEE Trans. Pattern Anal. Mach. Intell.}, vol. 39, no. 6, pp. 1137–1149, 2017.
\bibitem{ref8}
R. Krishna \emph{et al.}, “Visual genome: connecting language and vision using crowdsourced dense image annotations,” in \emph{Proc. Int. J. Comput. Vis.}, vol. 123, pp. 32–73, 2017.
\bibitem{ref9}
H. Jiang, I. Misra, M. Rohrbach, E. Learned-Miller, and X. Chen, “In defense of grid features for visual question answering,” in \emph{Proc. IEEE Conf. Comput. Vis. Pattern Recognit.}, 2020, pp.10267–10276.
\bibitem{ref10} 
Y. Liu, E. Sangineto, W. Bi, N. Sebe, B. Lepri, and M. Nadai, “Efficient training of visual transformers with small datasets”,  in \emph{Proc. Adv. Neural Inf. Process. Syst..}, Vol. 34, 2021, pp.23818–23830.
\bibitem{ref11} 
Z. Huang, Z. Zeng, Y. Huang, B. Liu, D. Fu, and J. Fu, “Seeing out of the box: end-to-end pre-training for vision-language representation learning,” in \emph{Proc. IEEE Conf. Comput. Vis. Pattern Recognit.}, pp.12976–12985, 2021.
\bibitem{ref12} 
W. Kim, B. Son, and I. Kim, “Vilt: vision-and-language transformer without convolution or region supervision”, in \emph{Proc. Int. Conf. Mach. Learn.}, pp. 5583–5594, 2021.
\bibitem{ref13} 
Z. Fang \emph{et al.}, “Injecting semantic concepts into end-to-end image captioning”, in \emph{Proc. IEEE Conf. Comput. Vis. Pattern Recognit.}, pp.18009–18019, 2022.
\bibitem{ref14} 
Y. Chen \emph{et al.}, “Uniter: universal image-text representation learning”, in \emph{Proc. IEEE Conf. Comput. Vis. Pattern Recognit.}, pp.104–120, 2020.
\bibitem{ref15} 
H. Bao \emph{et al.}, “Unilmv2: pseudo-masked language models for unified language model pre-training. in \emph{Proc. Int. Conf. Mach. Learn.}, pp. 642–652, 2020.
\bibitem{ref16} 
T. Lin \emph{et al.}, “Microsoft COCO: common objects in context”, in \emph{Proc. Eur. Conf. Comput. Vis.}, pp.740–755, 2014.
\bibitem{ref17} 
Z. Yang, X. He, J. Gao, L. Deng, and A. Smola, “Stacked attention networks for image question answering”, in \emph{Proc. IEEE Conf. Comput. Vis. Pattern Recognit.}, pp.21–29, 2016.
\bibitem{ref18} 
K. J. 	Shih, S. Singh, and D. Hoiem, “Where to look: focus regions for visual question answering”, in \emph{Proc. IEEE Conf. Comput. Vis. Pattern Recognit.}, pp.4613–4621, 2016.
\bibitem{ref19} 
Ye. Li, Y. Pan, T. Yao, and T. Mei, “Comprehending and ordering semantics for image captioning”, \emph{IEEE Trans. Circuits Syst. Video Technol.}, vol. 32, no. 10, pp.7005–7018, Oct. 2022.
\bibitem{ref20}
C. Yan \emph{et al.}, “Task-adaptive attention for image captioning”, \emph{IEEE Trans. Circuits Syst. Video Technol.}, vol. 32, no. 1, pp. 43–51, Jan. 2022.
\bibitem{ref21}
L. Zhou, H. Palangi, L. Zhang, H. Hu, J. J. Corso, and J. Gao, “Unified vision-language pretraining for image captioning and VQA”, in \emph{Proc. AAAI Conf. Artif. Intell.}, vol. 34, no. 7, pp.13041–13049, 2020. 
\bibitem{ref22} 
C. Li \emph{et al.}, “mPLUG: effective and efficient vision-language learning by cross-modal skip-connections”, 2022, \emph{arXiv:2205.12005}. [Online]. Available: http://arxiv.org/abs/2205.12005
\bibitem{ref23} 
 J. Lu \emph{et al.}, “12-in-1: multi-task vision and language representation learning”, in \emph{Proc. IEEE Conf. Comput. Vis. Pattern Recognit.}, pp.10437–10446, 2020.
\bibitem{ref24} 
Z. Huang, Z. Zeng, B. Liu, D. Fu, and J. Fu, “Pixel-BERT: aligning image pixels with text by deep multi-modal transformers”, 2020, \emph{arXiv:2004.00849}. [Online]. Available: http://arxiv.org/abs/2004.00849
\bibitem{ref25} 
S. Shen \emph{et al.}, “How much can clip benefit vision-and-language tasks?”, 2021, \emph{arXiv:2107.06383}. [Online]. Available: http://arxiv.org/abs/2107.06383
\bibitem{ref26} 
A. Dosovitskiy \emph{et al.}, “An image is worth 16x16 words: transformers for image recognition at scale”, in \emph{Proc. Int. Conf. Learn. Represent.}, 2021.
\bibitem{ref27} 
C. Lee, P. Panda, G. Srinivasan, and K. Roy, “Training deep spiking convolutional neural networks with stdp-based unsupervised pre-training followed by supervised fine-tuning”, \emph{Front. Neurosci.}, vol. 12, 2018.
\bibitem{ref28} 
W. Gerstner, W. M. Kistler, R. Naud, and L. Paninski, “Neuronal dynamics: from single neurons to networks and models of cognition”, \emph{Cambridge University Press}, 2014.
\bibitem{ref29} 
K. M. Stewart and E. O. Neftci, “Meta-learning spiking neural networks with surrogate gradient descent”, in \emph{Neuromorphic Computing and Engineering}, vol. 2, no. 4, 2022.
\bibitem{ref30} 
J. Li \emph{et al.}, “Align before fuse: vision and language representation learning with momentum distillation”, in \emph{Proc. Adv. Neural Inf. Process. Syst..}, vol. 34, pp. 9694–9705, 2021. 
\bibitem{ref31} 
H. Zheng, Y. Wu, L. Deng, Y. Hu, and G. Li, “Going deeper with directly-trained larger spiking neural networks”, in \emph{Proc. AAAI Conf. Artif. Intell.}, vol. 35, no. 12, pp. 11062–11070, 2021.
\bibitem{ref32} 
J. Devlin, M. Chang, K. Lee, and K. Toutanova, “Bert: pre-training of deep bidirectional transformers for language understanding”, in \emph{Proc. Annual Meeting of the ACL}, pp. 4171–4186, 2019.
\bibitem{ref33} 
K. He, H. Fan, Y. Wu, S. Xie, and R. Girshick, “Momentum contrast for unsupervised visual representation learning”, in \emph{Proc. IEEE Conf. Comput. Vis. Pattern Recognit.}, pp. 9729–9738, 2020.
\bibitem{ref34} 
J. Deng, W. Dong, R. Socher, L. Li, K. Li, and F. Li, “Imagenet: a large-scale hierarchical image database”, in \emph{Proc. IEEE Conf. Comput. Vis. Pattern Recognit.}, pp. 248–255, 2009.
\bibitem{ref35} 
Y. Wu \emph{et al.}, “Google’s neural machine translation system: bridging the gap between human and machine translation”, 2016, \emph{arXiv:1609.08144}. [Online]. Available: http://arxiv.org/abs/1609.08144
\bibitem{ref36} 
Y. Goyal\emph{et al.}, “Making the v in vqa matter: elevating the role of image understanding in visual question answering”, in \emph{Proc. IEEE Conf. Comput. Vis. Pattern Recognit.}, pp. 6904–6913, 2017.
\bibitem{ref37} 
M. Ren, R. Kiros, and R. Zemel, “Exploring models and data for image question answering”, in \emph{Proc. Adv. Neural Inf. Process. Syst..}, pp. 2953–2961, 2015.
\bibitem{ref38} 
A. Agrawal, D. Batra, D. Parikh, and A. Kembhavi, “Don’t just assume; look and answer: Overcoming priors for visual question answering”, in \emph{Proc. IEEE Conf. Comput. Vis. Pattern Recognit.}, pp. 4971–4980, 2018.
\bibitem{ref39} 
B. A. Plummer \emph{et al.}, “Flickr30k entities: collecting region-to-phrase correspondences for richer image-to-sentence models”, in \emph{Proc. IEEE Int. Conf. Comput. Vis.}, pp. 2641–2649, 2015.
\bibitem{ref40} 
A. Suhr, S. Zhou, A. Zhang, I. Zhang, H. Bai, and Y. Artzi, “A corpus for reasoning about natural language grounded in photographs”, in \emph{Proc. Annual Meeting of the ACL}, pp. 6418–6428, 2019.
\bibitem{ref41} 
N. Xie, F. Lai, D. Doran, and A. Kadav, “Visual entailment task for visually-grounded language learning,” in \emph{Proc. Adv. Neural Inf. Process. Syst.}, 2018.
\bibitem{ref42} 
L. Peng, Y. Yang, Z. Wang, X. Wu, and Z. Huang, “CRA-net: composed relation attention network for visual question answering”, in \emph{Proc. 27th ACM international conference on multimedia}, pp. 1202–1210, 2019. 
\bibitem{ref43} 
L. Li, Z. Gan, Y. Cheng, and J. Liu, “Relation-aware graph attention network for visual question answering,” in \emph{Proc. IEEE Int. Conf. Comput. Vis.}, pp. 10313–10322, 2019.
\bibitem{ref44} 
J. Lu, D. Batra, D. Parikh, and S. Lee, “Vilbert: pretraining task-agnostic visiolinguistic representations for vision-and-language tasks,” in \emph{Proc. Adv. Neural Inf. Process. Syst.}, pp. 13–23, 2019.
\bibitem{ref45} 
J. Wang, P. Zhou, M. Z. Shou, and S. Yan, “Position-guided text prompt for vision-language pre-training”, in \emph{Proc. IEEE Conf. Comput. Vis. Pattern Recognit.}, pp. 23242–23251, 2023.
\bibitem{ref46} 
H. Tan and M. Bansal, “LXMERT: Learning cross-modality encoder representations from transformers”, 2019, \emph{arXiv:1908.07490}. [Online]. Available: http://arxiv.org/abs/1908.07490
\bibitem{ref47} 
Y. Zhou \emph{et al.}, “TRAR: routing the attention spans in transformer for visual question answering,” in \emph{Proc. IEEE Int. Conf. Comput. Vis.}, pp. 2074–2084, 2021.
\bibitem{ref48} 
L. Peng, Y. Yang, Z. Wang, Z. Huang, and H. T. Shen, “MRA-Net: improving vqa via multi-modal relation attention network”, \emph{IEEE Trans. Pattern Anal. Mach. Intell.}, vol. 44, no. 1, pp. 318–329, 2020.
\bibitem{ref49} 
Z. Liang, H. Hu, and J. Zhu, “LPF: a language-prior feedback objective function for de-biased visual question answering,” in \emph{Proc. 44th ACM SIGIR Conference on Research and Development in Information Retrieval}, pp. 1955–1959, 2021.
\bibitem{ref50} 
X. Han, S. Wang, C. Su, Q. Huang, and Q. Tian, “Greedy gradient ensemble for robust visual question answering,” in \emph{Proc. IEEE Int. Conf. Comput. Vis.}, pp. 1584–1593, 2021.
\bibitem{ref51} 
Y. Zeng, X. Zhang, and H. Li, “Multi-grained vision language pre-training: aligning texts with visual concepts”, 2021, \emph{arXiv:2111.08276}. [Online]. Available: http://arxiv.org/abs/2111.08276
\bibitem{ref52} 
K. Lee, X. Chen, G. Hua, H. Hu, and X. He, “Stacked cross attention for image-text matching,” in \emph{Proc. Eur. Conf. Comput. Vis.}, pp. 201–216, 2018.
\bibitem{ref53} 
G. Li, N. Duan, Y. Fang, D. Jiang, and M. Zhou, “Unicoder-VL: a universal encoder for vision and language by cross-modal pre-training”, in \emph{Proc. AAAI Conf. Artif. Intell.}, pp. 11336–11344, 2019.
\bibitem{ref54} 
M. Cheng \emph{et al.}, “ViSTA: vision and scene text aggregation for cross-modal retrieval”, in \emph{Proc. IEEE Conf. Comput. Vis. Pattern Recognit.}, pp. 5184–5193, 2022.
\bibitem{ref55} 
J. Wang \emph{et al.}, “Omnivl: one foundation model for image-language and video-language tasks,” in \emph{Proc. Adv. Neural Inf. Process. Syst.}, vol. 35, pp. 5696–5710, 2022.
\bibitem{ref56} 
B. Jawade, D. D. Mohan, N. M. Ali, S. Setlur, and V. Govindaraju, “NAPReg: nouns as proxies regularization for semantically aware cross-modal embeddings,” in \emph{Proc. IEEE Winter Conference on Applications of Computer Vision}, pp. 1135–1144, 2023.
\bibitem{ref57} 
L. H. Li, M. Yatskar, D. Yin, C. Hsieh, and K. Chang, “VisualBERT: a simple and performant baseline for vision and language”, 2019, \emph{arXiv:1908.03557}. [Online]. Available: http://arxiv.org/abs/1908.03557
\bibitem{ref58} 
Z. Wang, J. Yu, A. W. Yu, Z. Dai, Y. Tsvetkov, and Y. Cao, “SimVLM: simple visual language model pretraining with weak supervision,” in \emph{Proc. Int. Conf. Learn. Represent.}, 2022.
\bibitem{ref59} 
H. Yang, J. Lin, A. Yang, P. Wang, C. Zhou, and H. Yang, “Prompt tuning for generative multimodal pretrained models”, 2022, \emph{arXiv:2208.02532}. [Online]. Available: http://arxiv.org/abs/2208.02532
\bibitem{ref60} 
J. Yu, Z. Wang, V. Vasudevan, L. Yeung, M. Seyedhosseini, and Y. Wu, “Coca: contrastive captioners are image-text foundation models, 2022, \emph{arXiv:2205.01917}. [Online]. Available: http://arxiv.org/abs/2205.01917
\end{thebibliography}
\end{document}